\documentclass[final,5p,times,twocolumn]{elsarticle}

\usepackage{amsmath}
\usepackage{amssymb}
\usepackage{color}
\usepackage{multirow}
\usepackage{array}
\usepackage{subfigure}
\usepackage{enumitem}
\usepackage{booktabs}
\usepackage{verbatim}
\biboptions{sort&compress}

\begin{document}

\begin{frontmatter}

\title{M$^{2}$RNet: Multi-modal and Multi-scale Refined Network for RGB-D Salient Object Detection}

\author[Address1]{Xian Fang}
\ead{xianfang@mail.nankai.edu.cn}
\author[Address2]{Jinchao Zhu}
\author[Address3]{Ruixun Zhang}
\author[Address1]{Xiuli Shao}
\author[Address2]{Hongpeng Wang\corref{*}}
\cortext[*]{Corresponding author.}
\ead{hpwang@nankai.edu.cn}
\address[Address1]{College of Computer Science, Nankai University, Tianjin 300350, China}
\address[Address2]{College of Artificial Intelligence, Nankai University, Tianjin 300350, China}
\address[Address3]{School of Mathematical Sciences, Peking University, Beijing 100871, China}

\begin{abstract}
  Salient object detection is a fundamental topic in computer vision. Previous methods based on RGB-D often suffer from the incompatibility of multi-modal feature fusion and the insufficiency of multi-scale feature aggregation. To tackle these two dilemmas, we propose a novel multi-modal and multi-scale refined network (M$^{2}$RNet). Three essential components are presented in this network. The nested dual attention module (NDAM) explicitly exploits the combined features of RGB and depth flows. The adjacent interactive aggregation module (AIAM) gradually integrates the neighbor features of high, middle and low levels. The joint hybrid optimization loss (JHOL) makes the predictions have a prominent outline. Extensive experiments demonstrate that our method outperforms other state-of-the-art approaches.
\end{abstract}

\begin{keyword}
Saliency detection \sep Deep learning \sep Multi-modal feature \sep Multi-scale feature \sep Loss function
\end{keyword}

\end{frontmatter}


\section{Introduction}

Salient object detection (SOD) aims to identify the most conspicuous object that attracts humans in the scene. It has been successfully applied in various fields, such as image retrieval \cite{Gao2015Database, Yang2015Scalable}, robot navigation \cite{Craye2016Environment}, person re-identification \cite{Zhao2013Unsupervised} and many more.

The SOD methods have exhibited broad prospects owing to the powerful representation ability of convolutional neural networks (CNNs) \cite{LeCun1998Gradient-based} and fully convolutional networks (FCNs) \cite{Long2015Fully}. Most of them resort to a single RGB information, which is difficult to achieve satisfactory results in complex scenes. At present, depth information has become growing popular thanks to the emergence of affordable and portable devices. As a supplement to RGB features, depth features provide rich distance information. However, the inherent differences between the multiple modalities lead to the bottleneck of feature fusion. Moreover, although the features of each scale have the detail or semantic information, it is hard to adequately aggregate them.

\begin{figure}[t] \small
  \centering
  \includegraphics[width=0.076\textwidth]{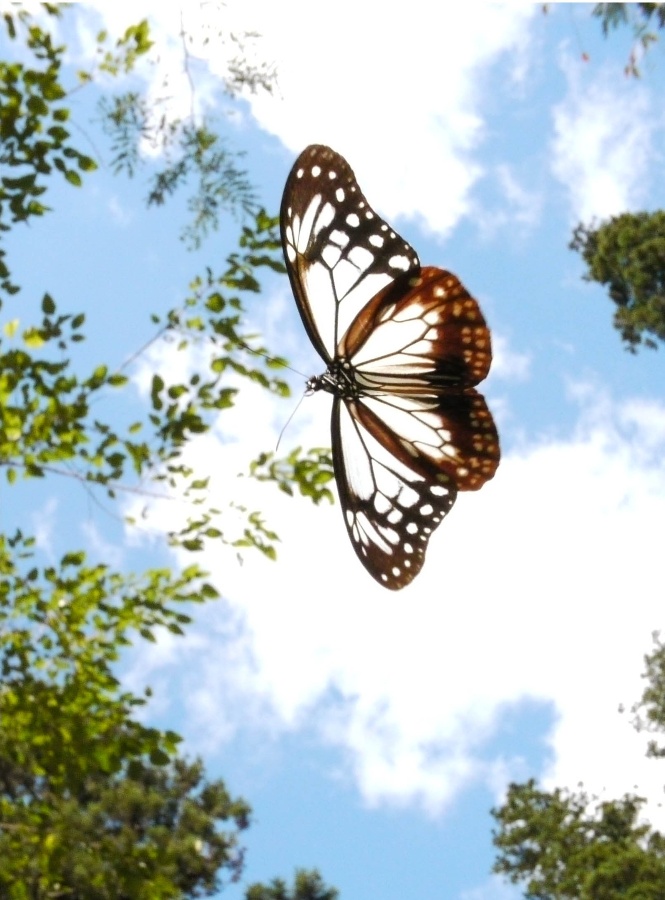}
  \includegraphics[width=0.076\textwidth]{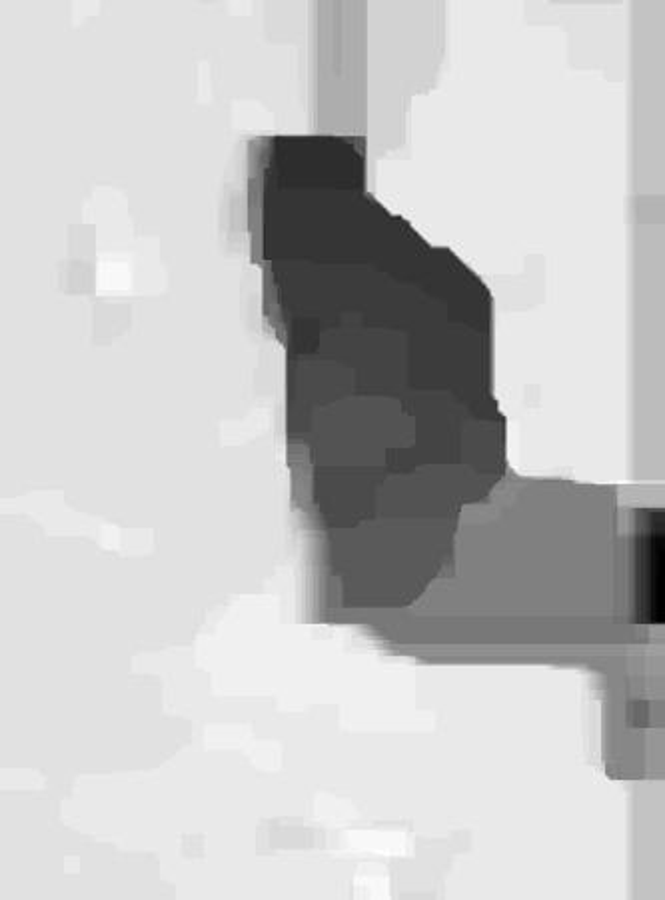}
  \includegraphics[width=0.076\textwidth]{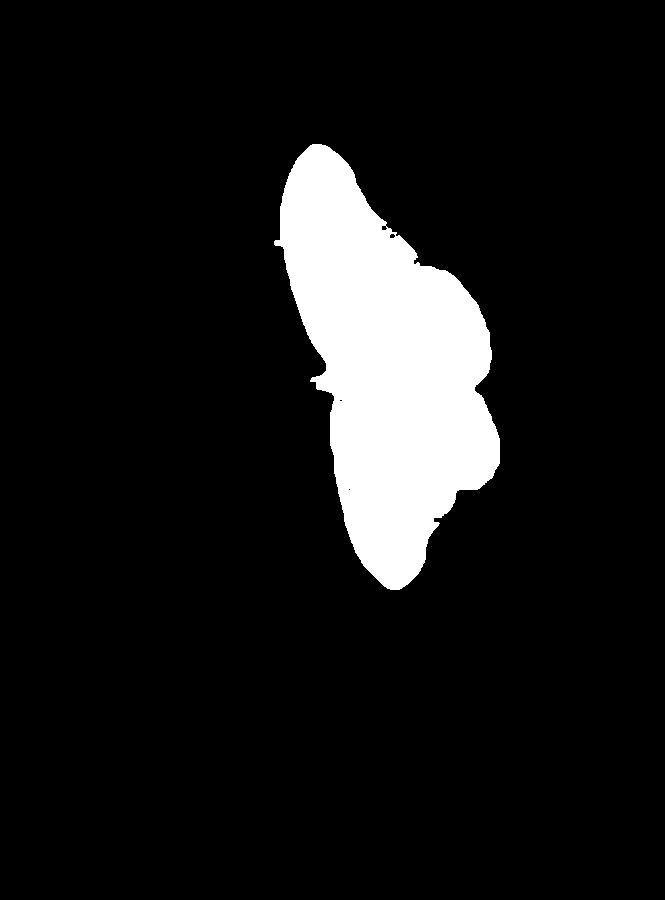}
  \includegraphics[width=0.076\textwidth]{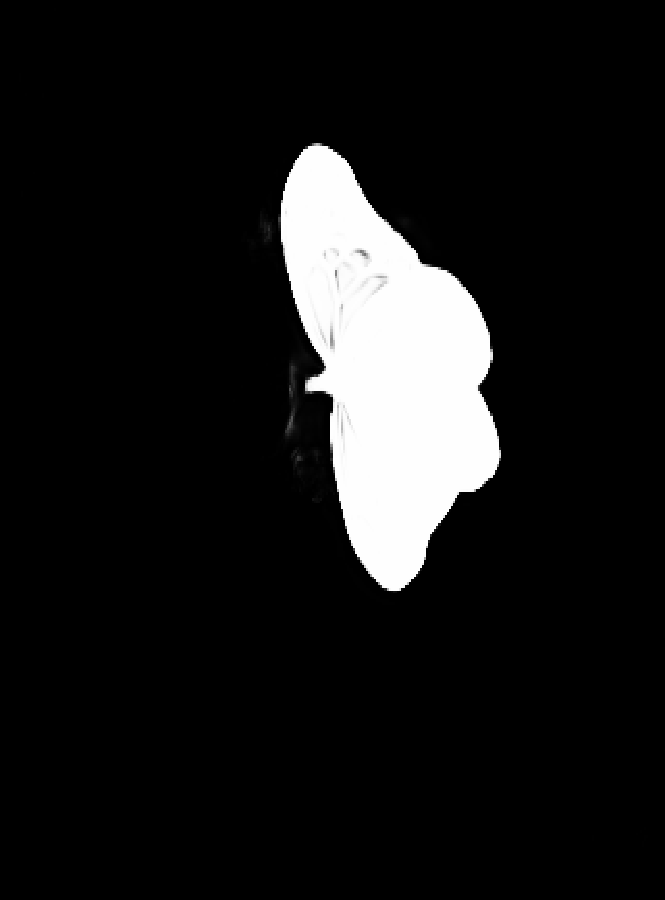}
  \includegraphics[width=0.076\textwidth]{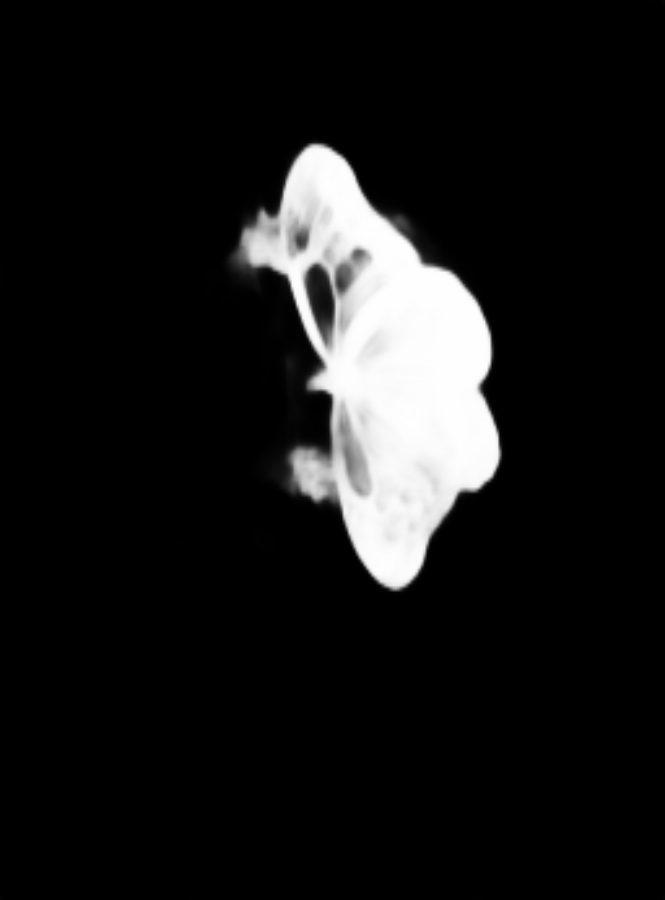}
  \includegraphics[width=0.076\textwidth]{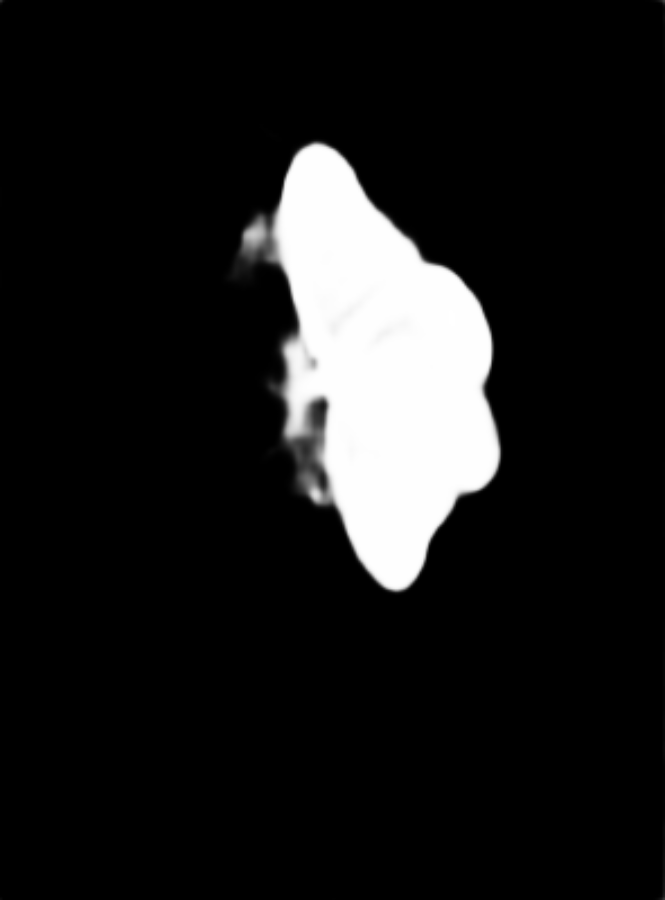} \\
  \vspace{-0.16cm}
  \subfigure[]{\includegraphics[width=0.076\textwidth]{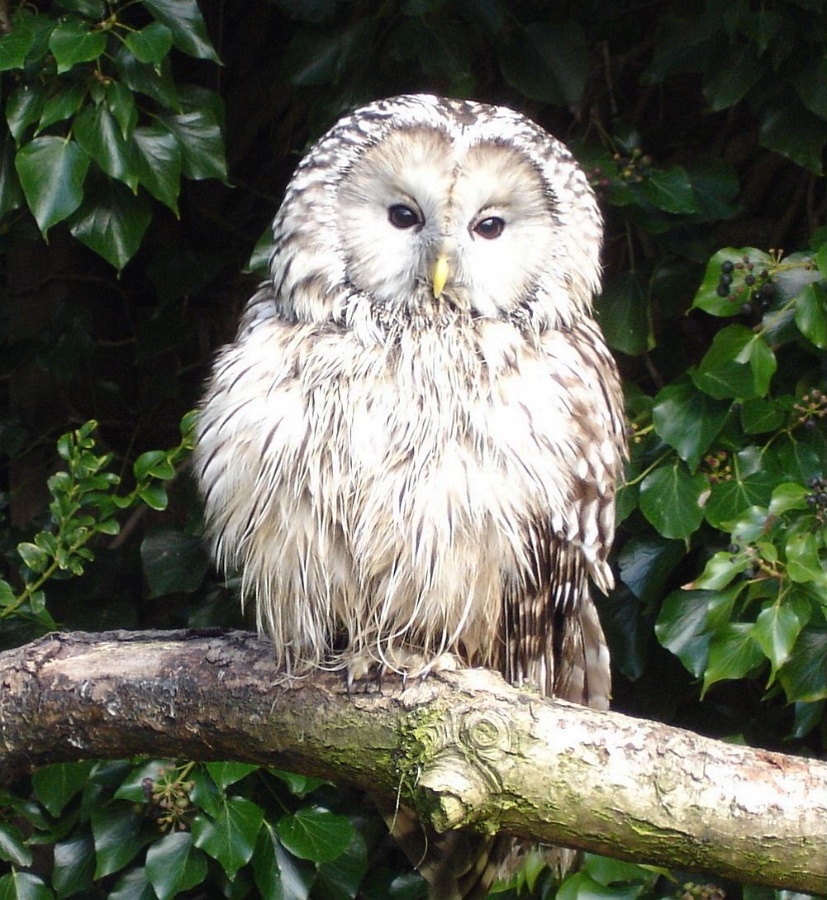}}
  \subfigure[]{\includegraphics[width=0.076\textwidth]{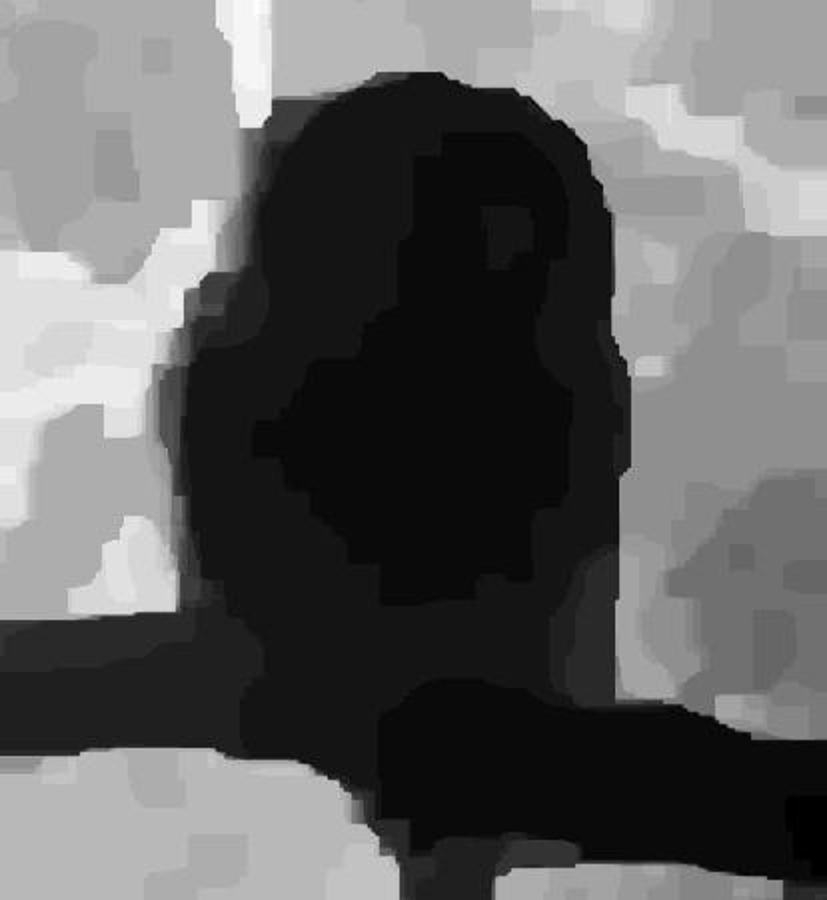}}
  \subfigure[]{\includegraphics[width=0.076\textwidth]{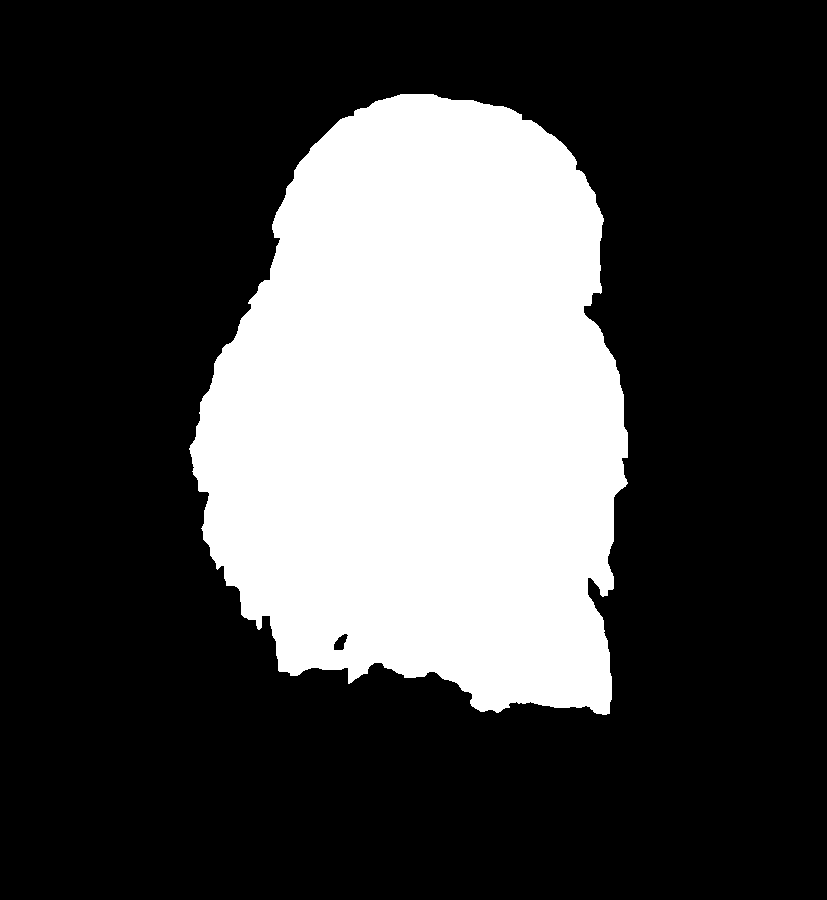}}
  \subfigure[]{\includegraphics[width=0.076\textwidth]{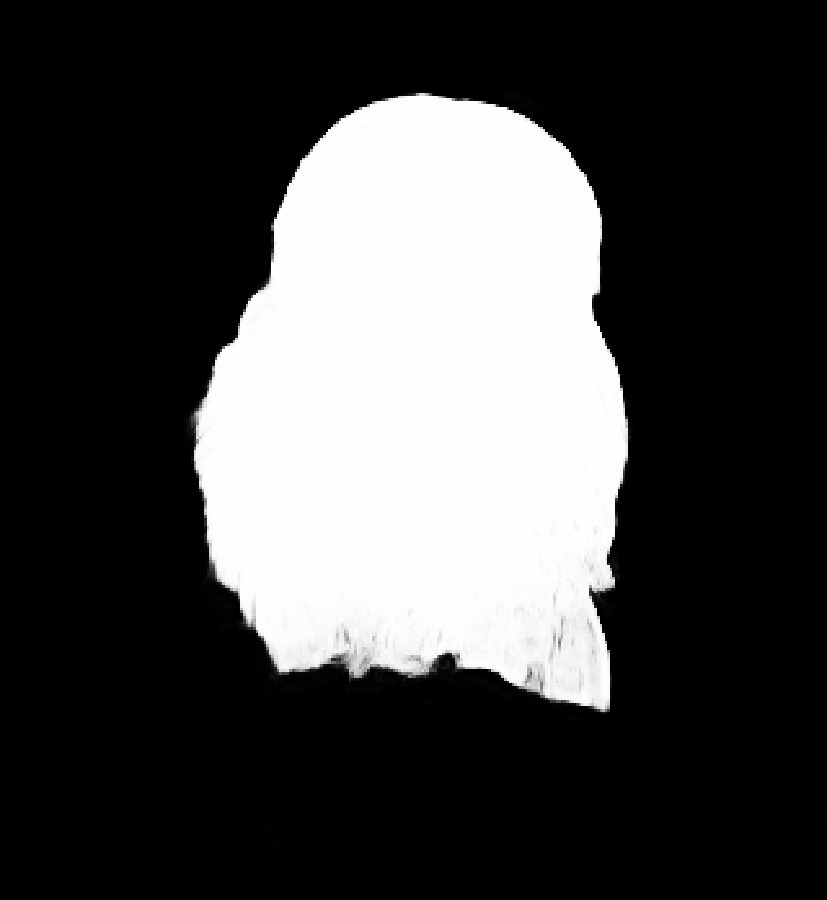}}
  \subfigure[]{\includegraphics[width=0.076\textwidth]{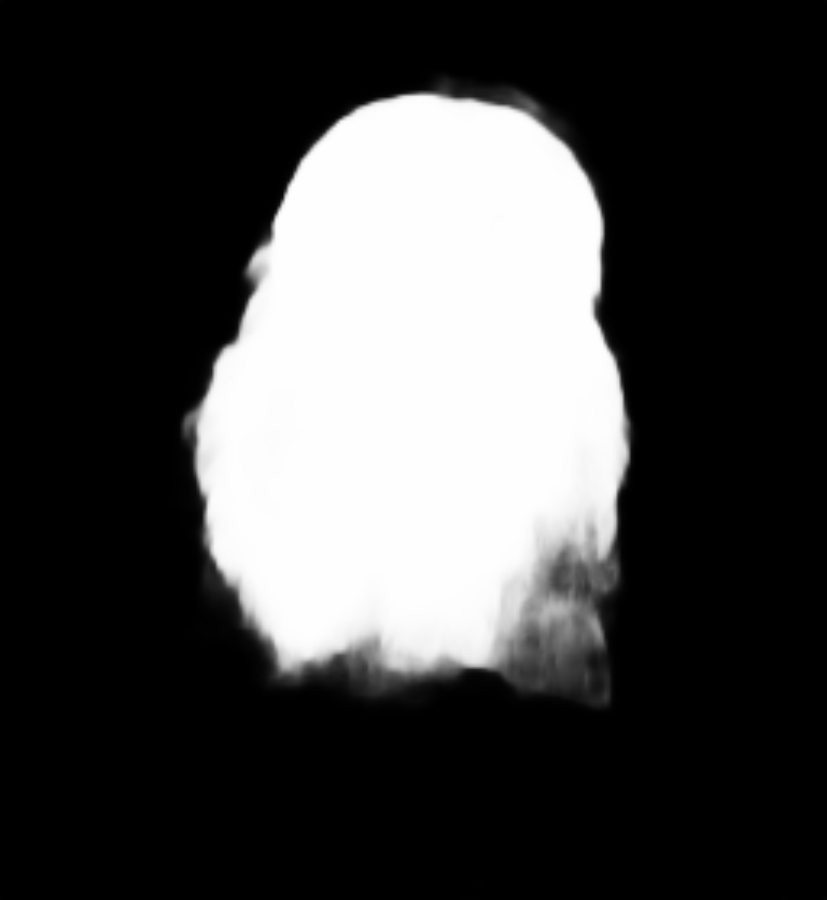}}
  \subfigure[]{\includegraphics[width=0.076\textwidth]{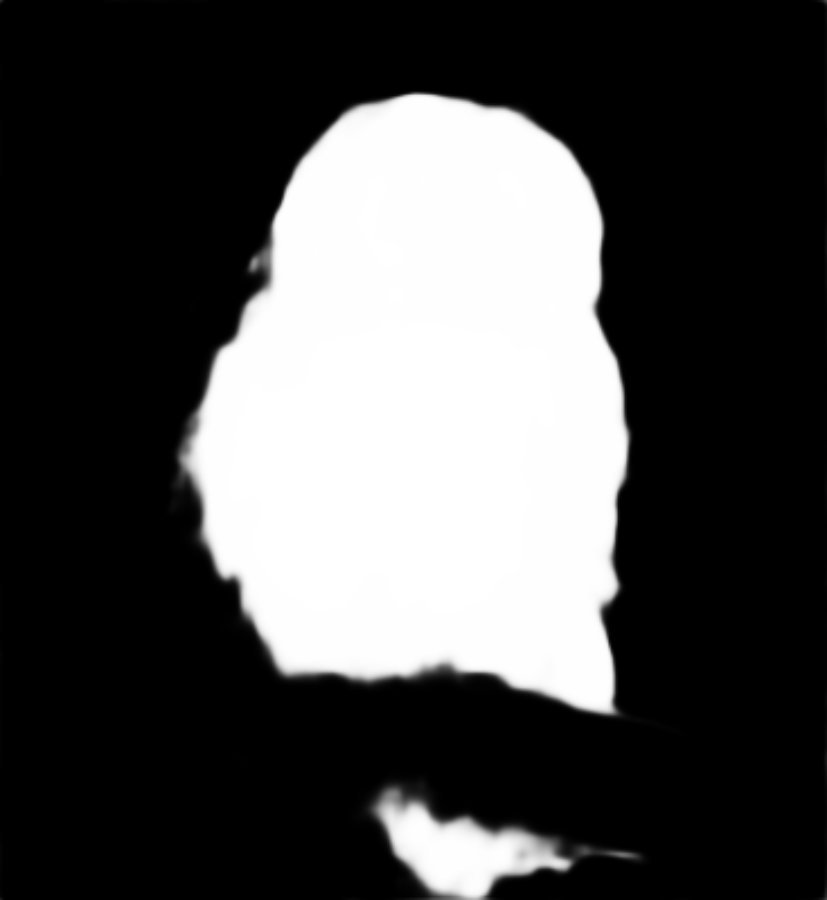}} \\
  \caption{Visual comparison of our method and two state-of-the-art approaches on several examples. (a) RGB; (b) Depth; (c) Ground truth; (d) Ours; (e) S$^{2}$MA \cite{Liu2020Learning}; (f) CPFP \cite{Zhao2019Contrast}.}
  \label{Figure 1}
\end{figure}

\begin{figure*}[t] \small
  \centering
  \includegraphics[width=0.99\textwidth]{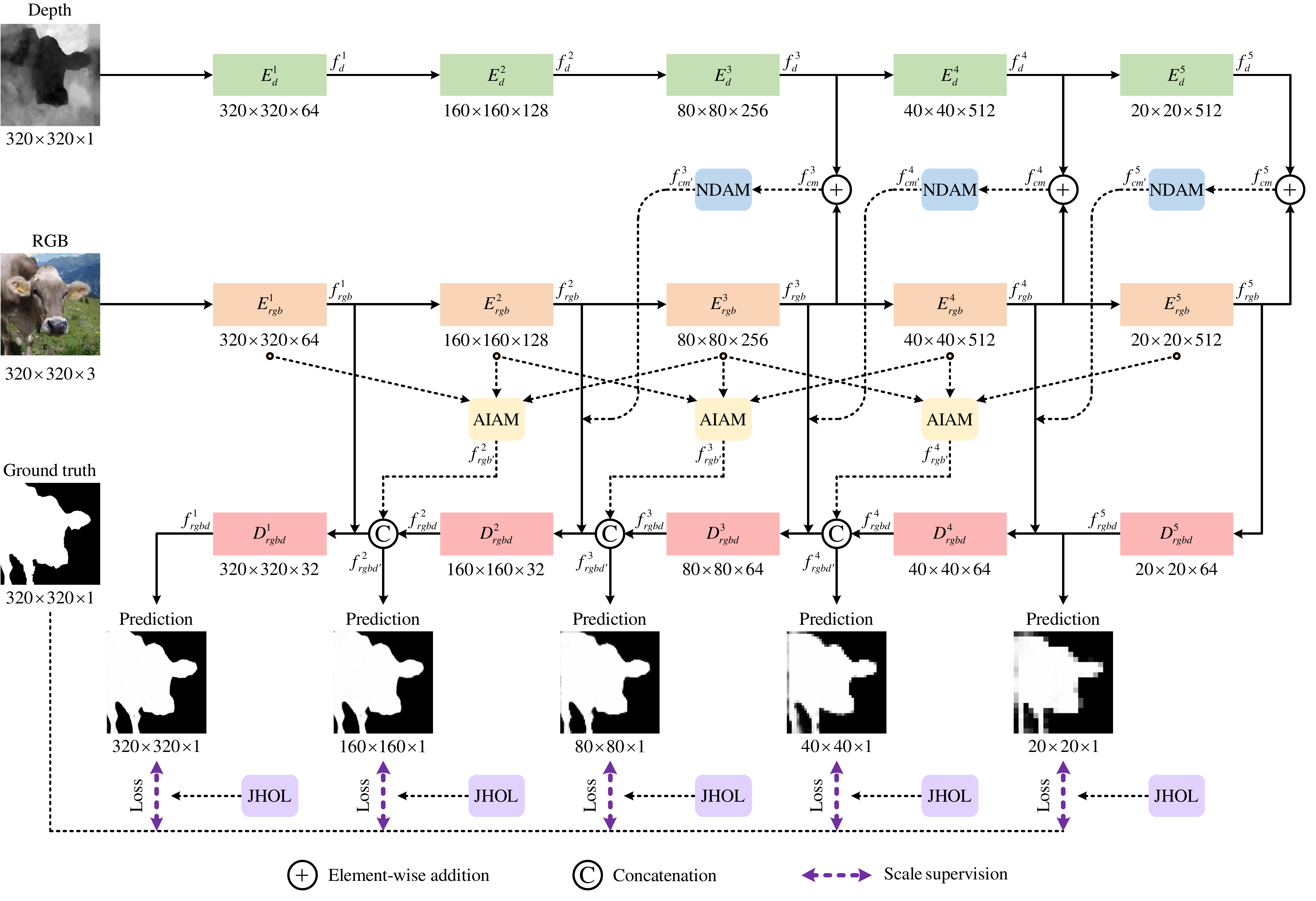} \\
  \caption{The overall architecture of the proposed M$^{2}$RNet. The backbone of our network is VGG-16 \cite{Simonyan2014Very}.}
  \label{Figure 2}
\end{figure*}

To this end, we propose a novel multi-modal and multi-scale refined network (M$^{2}$RNet) for RGB-D salient object detection. Specifically, the network is composed of the presented nested dual attention module (NDAM), adjacent interactive aggregation module (AIAM) and joint hybrid optimization loss (JHOL). In NDAM, we sequentially leverage channel and spatial attention to explicitly understand what and where is meaningful, thereby emphasizing or suppressing RGB and depth information that is important or unnecessary, for the purpose of boosting the fusion of RGB and depth features. In AIAM, we leverage the interaction of progressive and jumping connections in parallel to gradually learn information in abundant resolution for the purpose of boosting the aggregation of high level, middle level and low level features. In JHOL, we are devoted to guaranteeing inter-class discrimination and intra-class consistency by taking the local and global correlation of each pixel into account. These three components work together to achieve remarkable detection results. The visual comparison is shown in Fig. \ref{Figure 1}. Comparing with the saliency maps obtained by state-of-the-art approaches, those of our method are more exact.

Our main contributions are summarized as follows:
\begin{itemize}[noitemsep, nolistsep]
  \item We propose a multi-modal and multi-scale refined network (M$^{2}$RNet), which is equipped with the presented nested dual attention module (NDAM), adjacent interactive aggregation module (AIAM) and joint hybrid optimization loss (JHOL) components. Our network is capable of refining the multi-modal and multi-scale features simultaneously, nearly without extra computing cost under specific supervision.
  \item We conduct extensive experiments on seven datasets and demonstrate that our method achieves consistently superior performance against 12 state-of-the-art approaches including three RGB salient object detection approaches and nine RGB-D salient object detection approaches in terms of six evaluation metrics.
\end{itemize}

\section{Related Work}

\subsection{RGB Saliency Detection}

Lots of RGB salient object detection methods have been developed during the past decades.

For instance, Zhang et al. \cite{Zhang2018Progressive} proposed a PAGR, which selectively integrates multi-level contextual information. Liu et al. \cite{Liu2018PiCANet} proposed a PiCANet, which generates attention over the context regions for each pixel. Su et al. \cite{Su2019Selectivity} proposed a BANet, which enhances the feature selectivity at boundaries and keeps the feature invariance at interiors. Zhao et al. \cite{Zhao2019EGNet} proposed an EGNet, which explores edge information to preserve salient object boundaries. Wu et al. \cite{Wu2019Cascaded} proposed a CPD, which uses cascaded partial decoder to discards low-level features. Liu et al. \cite{Liu2019A} proposed a PoolNet, which explores the potentials of pooling. Zhang et al. \cite{Zhang2019CapSal} proposed a CapSal, which uses image captioning for detecting. Qin et al. \cite{Qin2019BASNet} proposed a BASNet, which focuses on end-to-end boundary-aware. Pang et al. \cite{Pang2020Multi-scale} proposed a MINet, which exchanges information between multi-scale. Wei et al. \cite{Wei2020Label} proposed a LDF, which decouples the saliency label into body map and detail map for iterative information exchange.

\subsection{RGB-D Saliency Detection}

Most recently, RGB-D salient object detection methods have rapidly aroused the concern of researchers and made impressive progress.

For instance, Chen et al. \cite{Chen2019Three-stream} proposed a TANet, which combines the bottom-up stream and the top-down stream to learn cross-modal complementarity. Wang et al. \cite{Wang2019Adaptive} proposed an AFNet, which adaptively fuses the predictions from the separate RGB and depth streams using the switch map. Piao et al. \cite{Piao2019Depth-induced} proposed a DMRA, which involves residual connections, multi-scale weighting and recurrent attention. Zhao et al. \cite{Zhao2019Contrast} proposed a CPFP by making use of feature fusion of contrast prior and fluid pyramid. Fan et al. \cite{Fan2019Rethinking} proposed a D$^3$Net, which automatically discards the low-quality depth maps via gate connection. Li et al. \cite{Li2020ICNet} proposed an ICNet, which can learn the optimal conversion of RGB features and depth features to autonomously merge them. Pang et al. \cite{Pang2020Hierarchical} proposed a HDFNet, in which the features of the network are densely connected and through the dynamic expansion pyramid. Fan et al. \cite{Fan2020BBS-Net} proposed a BBS-Net, in which multi-level features are partitioned into teacher and student features in the cascade network. Zhao et al. \cite{Zhao2020A} proposed a DANet, which explores early fusion and middle fusion between RGB and depth. Zhao et al. \cite{Zhao2020Cross-modal} proposed a CMWNet, which weights the fusion of low, medium and high levels to encourage feature interaction. Fu et al. \cite{Fu2020JL-DCF} proposed a JL-DCF for joint learning and densely-cooperative fusion. Zhang et al. \cite{Zhang2020UC-Net} proposed an UC-Net, which learns the distribution of saliency maps by conditional variational autoencoder. Piao et al. \cite{Piao2020A2dele} proposed an A2dele, which uses network prediction and attention as two bridges to transfer deep knowledge from deep stream to RGB stream. Liu et al. \cite{Liu2020Learning} proposed a S$^2$MA, which reweights the mutual attention for filtering out unreliable modality information.

\section{Proposed Method}

\subsection{Network Overview}

The proposed multi-modal and multi-scale refined network (M$^{2}$RNet) is an encoder-decoder architecture, covering nested dual attention module (NDAM), adjacent interactive aggregation module (AIAM) and joint hybrid optimization loss (JHOL), as shown in Fig. \ref{Figure 2}. To be concise, we denote the output features of RGB branch in the encoder as $f_{d}^{i}$ $(i = 1, 2, 3, 4, 5)$, the output features of depth branch in the encoder as $f_{rgb}^{i}$ $(i = 1, 2, 3, 4, 5)$, and the output features in the decoder as $f_{rgbd}^{i}$ $(i = 1, 2, 3, 4, 5)$.  Let $f_{cm}^{i}$ $(i = 3, 4, 5)$ denote the combined features of RGB flow and depth flow, we utilize NDAM to strengthen their robustness to get the corresponding enhanced features $f_{cm^{'}}^{i}$ $(i = 3, 4, 5)$. For each group of three consecutive features \{$f_{rgb}^{i-1}$, $f_{rgb}^{i}$, $f_{rgb}^{i+1}$\} $(i = 2, 3, 4)$, we utilize AIAM to produce their aggregated features $f_{rgb^{'}}^{i}$ $(i = 2, 3, 4)$. In order to facilitate the optimization, we embed JHOL as the auxiliary loss.

\begin{figure}[t]
  \centering
  \includegraphics[width=0.45\textwidth]{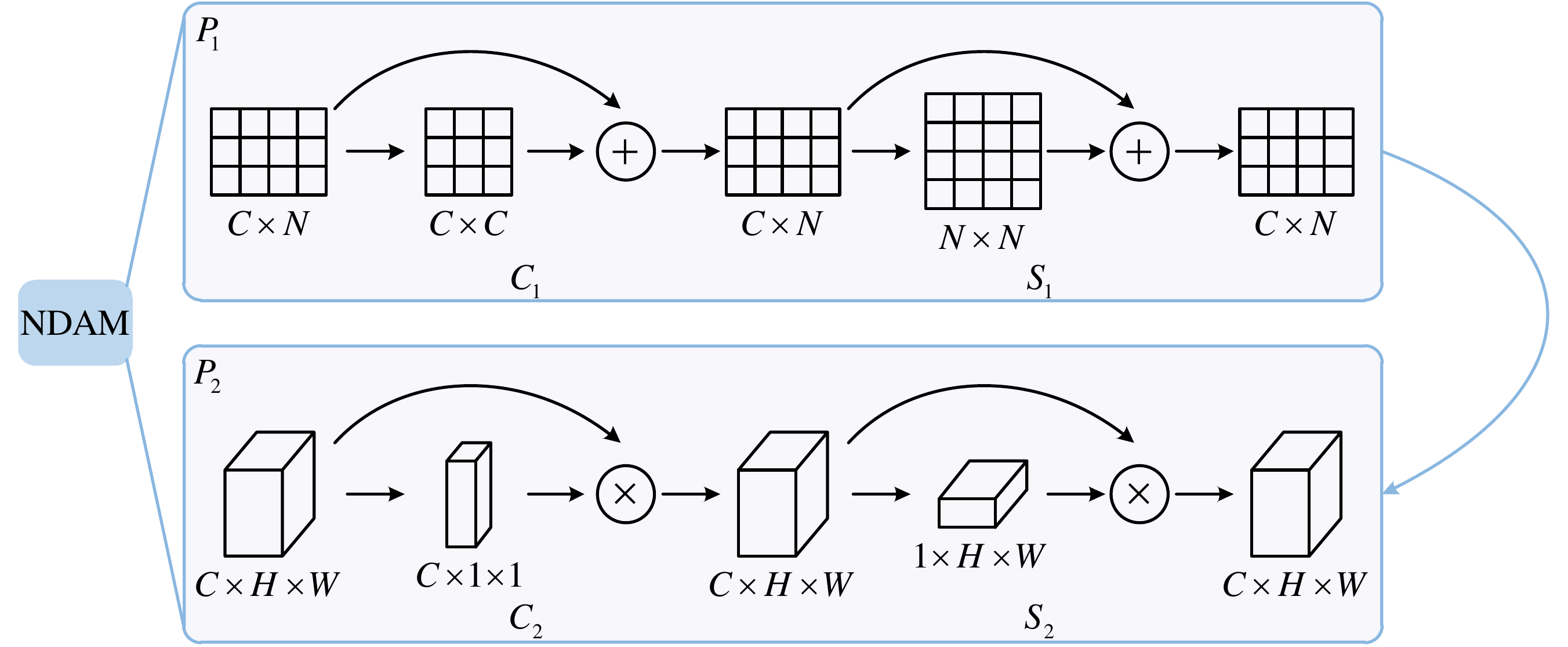} \\
  \caption{Illustration of NDAM.}
  \label{Figure 3}
\end{figure}

\begin{figure}[t]
  \centering
  \includegraphics[width=0.45\textwidth]{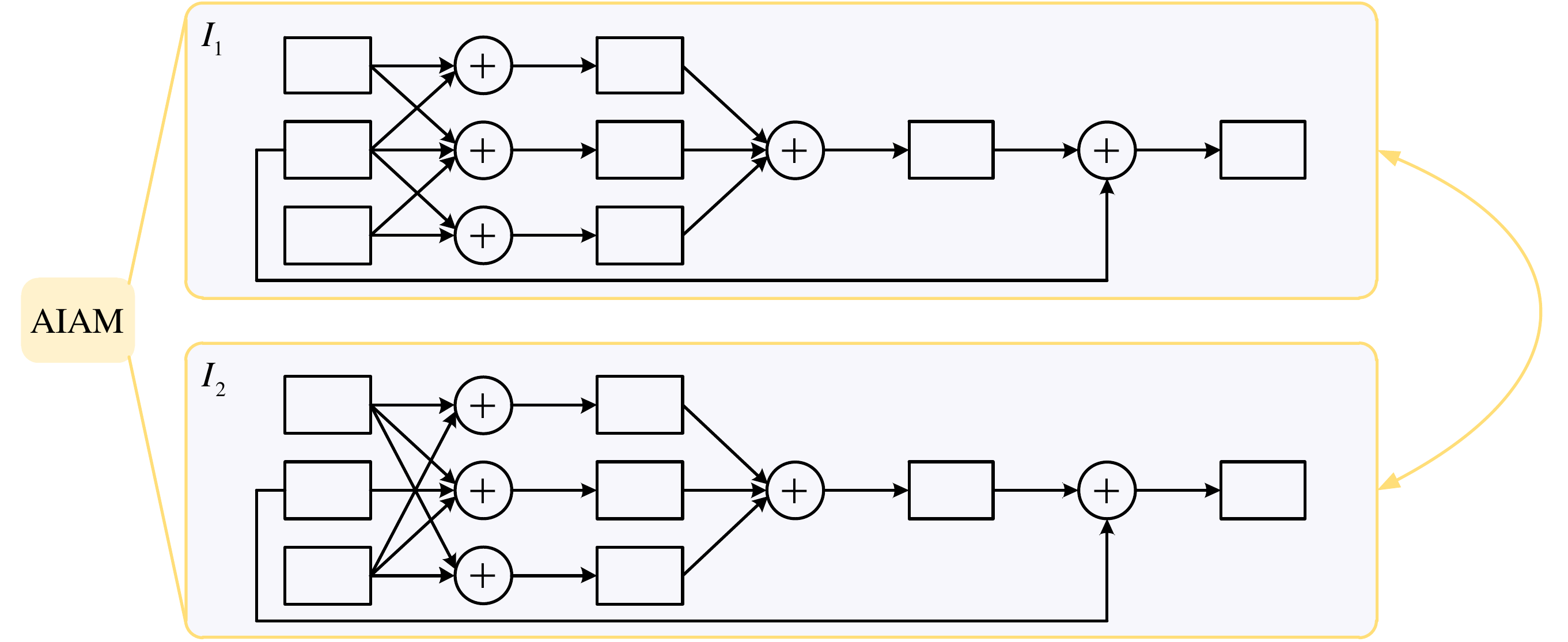} \\
  \caption{Illustration of AIAM.}
  \label{Figure 4}
\end{figure}

\begin{table*}[!t] 
  \caption{Quantitative comparison of different methods on six datasets in terms of six evaluation metrics. $\uparrow$ and $\downarrow$ indicate that the larger and smaller scores are better, respectively. The best three results are highlight in \textcolor{red}{red}, \textcolor{blue}{blue} and \textcolor{green}{green}. $\dag$ and $\ddag$ mean using VGG-19 \cite{Simonyan2014Very} and ResNet-50 \cite{He2016Deep} as the backbone, respectively. If not marked, VGG-16 \cite{Simonyan2014Very} is used as the backbone. '--' means no data available.}
  \label{Table 1}
  \centering
  \vspace{0.1cm}
  \resizebox{\textwidth}{!}{
    \begin{tabular}{cc|ccc|ccccccccc|c}
      \hline
      \multirow{2}{*}{Dataset} &Evaluation &EGNet &CPD &PoolNet &DF &CTMF &MMCI &TANet &AFNet &DMRA$^{\dag}$ &CPFP &D$^{3}$Net$^{\ddag}$ &S$^{2}$MA &M$^{2}$RNet \\
                               &metric &\cite{Zhao2019EGNet} &\cite{Wu2019Cascaded} &\cite{Liu2019A} &\cite{Qu2017RGBD} &\cite{Han2017CNNs-based} &\cite{Chen2019Multi-modal} &\cite{Chen2019Three-stream} &\cite{Wang2019Adaptive} &\cite{Piao2019Depth-induced} &\cite{Zhao2019Contrast} &\cite{Fan2019Rethinking} &\cite{Liu2020Learning} &(ours) \\
      \hline
      \multirow{4}{*}{STEREO \cite{Niu2012Leveraging}}
      &$S_{\alpha} \uparrow$         &-- &-- &-- &0.757 &0.848 &0.873 &0.871 &0.825 &0.752 &0.879 &\textcolor{blue}{0.891} &\textcolor{green}{0.890} &\textcolor{red}{0.899} \\
      &$F^{max}_{\beta} \uparrow$    &-- &-- &-- &0.789 &0.848 &0.877 &0.878 &0.848 &0.802 &0.889 &\textcolor{blue}{0.897} &\textcolor{green}{0.895} &\textcolor{red}{0.913} \\
      &$F^{avg}_{\beta} \uparrow$    &-- &-- &-- &0.742 &0.771 &0.829 &\textcolor{green}{0.835} &0.807 &0.762 &0.830 &0.833 &\textcolor{blue}{0.855} &\textcolor{red}{0.867} \\
      &$F^{\omega}_{\beta} \uparrow$ &-- &-- &-- &0.549 &0.698 &0.760 &0.787 &0.752 &0.647 &\textcolor{green}{0.817} &0.815 &\textcolor{blue}{0.825} &\textcolor{red}{0.851} \\
      &$E_{\xi} \uparrow$            &-- &-- &-- &0.838 &0.870 &0.905 &\textcolor{green}{0.916} &0.887 &0.816 &0.907 &0.911 &\textcolor{blue}{0.926} &\textcolor{red}{0.929} \\
      &$\mathcal{M} \downarrow$      &-- &-- &-- &0.141 &0.086 &0.068 &0.060 &0.075 &0.086 &\textcolor{blue}{0.051} &\textcolor{green}{0.054} &\textcolor{blue}{0.051} &\textcolor{red}{0.042} \\
      \hline
      \multirow{4}{*}{NLPR \cite{Peng2014RGBD}}
      &$S_{\alpha} \uparrow$         &0.867 &0.885 &0.867 &0.769 &0.860 &0.856 &0.886 &0.798 &0.899 &0.888 &\textcolor{green}{0.905} &\textcolor{blue}{0.915} &\textcolor{red}{0.918} \\
      &$F^{max}_{\beta} \uparrow$
      &0.857 &0.889 &0.844 &0.753 &0.840 &0.841 &0.876 &0.816 &0.888 &0.888 &\textcolor{green}{0.905} &\textcolor{blue}{0.910} &\textcolor{red}{0.921} \\
      &$F^{avg}_{\beta} \uparrow$    &0.800 &0.840 &0.791 &0.682 &0.723 &0.729 &0.795 &0.746 &\textcolor{blue}{0.855} &0.821 &0.832 &\textcolor{green}{0.846} &\textcolor{red}{0.862} \\
      &$F^{\omega}_{\beta} \uparrow$ &0.774 &0.829 &0.771 &0.524 &0.691 &0.688 &0.789 &0.699 &\textcolor{green}{0.846} &0.819 &0.833 &\textcolor{red}{0.855} &\textcolor{blue}{0.848} \\
      &$E_{\xi} \uparrow$            &0.910 &0.925 &0.900 &0.838 &0.869 &0.871 &0.916 &0.884 &\textcolor{red}{0.942} &0.923 &0.932 &\textcolor{green}{0.937} &\textcolor{blue}{0.941} \\
      &$\mathcal{M} \downarrow$      &
      0.047 &0.037 &0.046 &0.099 &0.056 &0.856 &0.041 &0.060 &\textcolor{blue}{0.031} &0.036 &0.034 &\textcolor{red}{0.030} &\textcolor{green}{0.033} \\
      \hline
      \multirow{4}{*}{RGBD135 \cite{Cheng2014Depth}}
      &$S_{\alpha} \uparrow$ &0.876 &0.891 &0.886 &0.681 &0.863 &0.848 &0.858 &0.770 &0.900 &0.872 &\textcolor{green}{0.904} &\textcolor{red}{0.941} &\textcolor{blue}{0.934} \\
      &$F^{max}_{\beta} \uparrow$    &0.900 &0.910 &0.906 &0.626 &0.865 &0.839 &0.853 &0.775 &0.907 &0.882 &\textcolor{green}{0.917} &\textcolor{red}{0.944} &\textcolor{blue}{0.937} \\
      &$F^{avg}_{\beta} \uparrow$    &0.843 &0.869 &0.864 &0.573 &0.778 &0.762 &0.795 &0.730 &0.866 &0.829 &\textcolor{green}{0.876} &\textcolor{blue}{0.906} &\textcolor{red}{0.910} \\
      &$F^{\omega}_{\beta} \uparrow$ &0.780 &0.824 &0.807 &0.383 &0.686 &0.650 &0.739 &0.641 &\textcolor{green}{0.843} &0.787 &0.831 &\textcolor{blue}{0.892} &\textcolor{red}{0.903} \\
      &$E_{\xi} \uparrow$            &0.930 &0.930 &0.940 &0.806 &0.911 &0.904 &0.919 &0.874 &0.944 &0.927 &\textcolor{green}{0.956} &\textcolor{red}{0.974} &\textcolor{blue}{0.971} \\
      &$\mathcal{M} \downarrow$      &0.037 &0.032 &0.032 &0.132 &0.055 &0.065 &0.046 &0.068 &\textcolor{green}{0.030} &0.038 &\textcolor{green}{0.030} &\textcolor{blue}{0.021} &\textcolor{red}{0.019} \\
      \hline
      \multirow{4}{*}{LFSD \cite{Li2014Saliency}}
      &$S_{\alpha} \uparrow$         &0.818 &0.806 &0.826 &0.776 &0.796 &0.787 &0.801 &0.738 &\textcolor{red}{0.847} &0.828 &0.832 &\textcolor{green}{0.837} &\textcolor{blue}{0.842} \\
      &$F^{max}_{\beta} \uparrow$    &0.838 &0.834 &0.846 &0.854 &0.815 &0.813 &0.827 &0.780 &\textcolor{red}{0.872} &0.850 &0.849 &\textcolor{blue}{0.862} &\textcolor{green}{0.861} \\
      &$F^{avg}_{\beta} \uparrow$    &0.803 &0.808 &0.790 &0.811 &0.780 &0.779 &0.786 &0.742 &\textcolor{red}{0.849} &0.813 &0.801 &\textcolor{green}{0.820} &\textcolor{blue}{0.825} \\
      &$F^{\omega}_{\beta} \uparrow$ &0.745 &0.753 &0.757 &0.618 &0.695 &0.663 &0.718 &0.671 &\textcolor{red}{0.811} &\textcolor{green}{0.775} &0.756 &0.772 &\textcolor{blue}{0.786} \\
      &$E_{\xi} \uparrow$            &0.854 &0.856 &0.852 &0.841 &0.851 &0.840 &0.845 &0.810 &\textcolor{red}{0.899} &0.867 &0.860 &\textcolor{blue}{0.876} &\textcolor{green}{0.874} \\
      &$\mathcal{M} \downarrow$      &0.102 &0.097 &\textcolor{green}{0.094} &0.151 &0.120 &0.132 &0.111 &0.133 &\textcolor{red}{0.075} &\textcolor{blue}{0.088} &0.099 &\textcolor{green}{0.094} &\textcolor{blue}{0.088} \\
      \hline
      \multirow{4}{*}{NJU2K \cite{Ju2015Depth-aware}}
      &$S_{\alpha} \uparrow$         &0.869 &0.862 &0.872 &0.735 &0.849 &0.859 &0.878 &0.771 &\textcolor{green}{0.886} &-- &\textcolor{blue}{0.895} &-- &\textcolor{red}{0.910} \\
      &$F^{max}_{\beta} \uparrow$    &0.880 &0.880 &0.887 &0.790 &0.857 &0.868 &0.888 &0.804 &\textcolor{green}{0.896} &-- &\textcolor{blue}{0.903} &-- &\textcolor{red}{0.922} \\
      &$F^{avg}_{\beta} \uparrow$    &\textcolor{green}{0.846} &\textcolor{blue}{0.853} &0.850 &0.744 &0.779 &0.803 &0.844 &0.766 &\textcolor{red}{0.872} &-- &0.819 &-- &0.841 \\
      &$F^{\omega}_{\beta} \uparrow$ &0.808 &0.821 &0.816 &0.553 &0.731 &0.749 &0.812 &0.699 &\textcolor{blue}{0.853} &-- &\textcolor{green}{0.839} &-- &\textcolor{red}{0.854} \\
      &$E_{\xi} \uparrow$            &0.905 &\textcolor{green}{0.908} &\textcolor{green}{0.908} &0.818 &0.864 &0.878 &\textcolor{blue}{0.909} &0.846 &\textcolor{red}{0.921} &-- &0.891 &-- &0.904 \\
      &$\mathcal{M} \downarrow$      &0.060 &0.059 &\textcolor{green}{0.057} &0.151 &0.085 &0.079 &0.061 &0.103 &\textcolor{blue}{0.051} &-- &\textcolor{blue}{0.051} &-- &\textcolor{red}{0.049} \\
      \hline
      \multirow{4}{*}{DUT-RGBD \cite{Piao2019Depth-induced}}
      &$S_{\alpha} \uparrow$         &0.872 &0.874 &\textcolor{blue}{0.892} &0.719 &0.830 &0.791 &0.808 &-- &\textcolor{green}{0.888} &0.749 &-- &\textcolor{red}{0.903} &\textcolor{red}{0.903} \\
      &$F^{max}_{\beta} \uparrow$    &0.897 &0.892 &0.907 &0.775 &0.842 &0.804 &0.823 &-- &\textcolor{green}{0.908} &0.787 &-- &\textcolor{blue}{0.909} &\textcolor{red}{0.925} \\
      &$F^{avg}_{\beta} \uparrow$    &0.861 &0.863 &\textcolor{green}{0.866} &0.748 &0.790 &0.751 &0.771 &-- &\textcolor{blue}{0.883} &0.735 &-- &\textcolor{green}{0.866} &\textcolor{red}{0.892} \\
      &$F^{\omega}_{\beta} \uparrow$ &0.797 &0.819 &0.829 &0.514 &0.681 &0.626 &0.703 &-- &\textcolor{green}{0.852} &0.636 &-- &\textcolor{blue}{0.856} &\textcolor{red}{0.864} \\
      &$E_{\xi} \uparrow$            &0.914 &0.915 &\textcolor{green}{0.924} &0.842 &0.882 &0.855 &0.866 &-- &\textcolor{blue}{0.930} &0.815 &-- &0.921 &\textcolor{red}{0.935} \\
      &$\mathcal{M} \downarrow$      &0.060 &0.059 &0.050 &0.150 &0.097 &0.113 &0.093 &-- &\textcolor{green}{0.048} &0.100 &-- &\textcolor{blue}{0.046} &\textcolor{red}{0.042} \\
      \hline
      \multirow{4}{*}{SIP \cite{Fan2019Rethinking}}
      &$S_{\alpha} \uparrow$         &-- &-- &-- &0.653 &0.716 &0.833 &0.835 &0.720 &0.800 &\textcolor{green}{0.850} &\textcolor{blue}{0.864} &-- &\textcolor{red}{0.882} \\
      &$F^{max}_{\beta} \uparrow$    &-- &-- &-- &0.704 &0.720 &0.840 &0.851 &0.756 &0.847 &\textcolor{green}{0.870} &\textcolor{blue}{0.882} &-- &\textcolor{red}{0.902} \\
      &$F^{avg}_{\beta} \uparrow$    &-- &-- &-- &0.673 &0.684 &0.795 &0.809 &0.705 &0.815 &\textcolor{green}{0.819} &\textcolor{blue}{0.831} &-- &\textcolor{red}{0.868} \\
      &$F^{\omega}_{\beta} \uparrow$ &-- &-- &-- &0.406 &0.535 &0.712 &0.748 &0.617 &0.734 &\textcolor{green}{0.788} &\textcolor{blue}{0.793} &-- &\textcolor{red}{0.840} \\
      &$E_{\xi} \uparrow$            &-- &-- &-- &0.794 &0.824 &0.886 &0.894 &0.815 &0.858 &\textcolor{green}{0.899} &\textcolor{blue}{0.903} &-- &\textcolor{red}{0.921} \\
      &$\mathcal{M} \downarrow$      &-- &-- &-- &0.185 &0.139 &0.086 &0.075 &0.118 &0.088 &\textcolor{green}{0.064} &\textcolor{blue}{0.063} &-- &\textcolor{red}{0.049} \\
      \hline
    \end{tabular}}
  \label{bigtable}
\end{table*}

\subsection{Nested Dual Attention Module}

There are some main issues in fusing RGB and depth features. The key point is that RGB and depth features are incompatible to a certain extent, which is due to the inherent differences between the two modalities. Besides, low-quality depth maps also inevitably bring more noise than cues.

In view of these, we propose a nested dual attention module (NDAM) to promote the coordination of multi-modal features and reduce the noise contamination of depth map. The two-phase attention mechanism is elaborately designed to mine potential features. Here, the channel attention mechanism of each phase is responsible for excavating the inter-channel relationship of features, while the spatial attention mechanism of each phase is responsible for excavating the inter-spatial relationship of features. By directly merging the RGB features $f_{rgb}^{i}$ $(i = 3, 4, 5)$ and the depth features $f_{d}^{i}$ $(i = 3, 4, 5)$, the combined features $f_{cm}^{i}$ $(i = 3, 4, 5)$ are easily calculated. Furthermore, the corresponding enhanced features $f_{cm^{'}}^{i}$ $(i = 3, 4, 5)$ are eventually obtained after the reinforcement of nested attention. The procedure of RGB and depth feature fusion can be described as:
\begin{equation}
  f_{cm}^{i} = f_{rgb}^{i} + f_{d}^{i},
  \label{Equation 1}
\end{equation}
\begin{equation}
  f_{cm^{'}}^{i} = S_{2}(C_{2}(S_{1}(C_{1}(f_{cm}^{i}))),
  \label{Equation 2}
\end{equation}
where $C_{i}(\cdot)$ and $S_{i}(\cdot)$ $(i = 1, 2)$ denote the channel attention and spatial attention, respectively. The nested attention is roughly divided into two phases of $P_{1}$ and $P_{2}$, as shown in Fig. \ref{Figure 3}.

In the first phase $P_{1}$, given an intermediate feature $\overline{f}\in \mathbb{R}^{C\times X}$, in which $X=H\times W$, and $C$, $H$ and $W$ are the channel, height and width of the feature $\overline{f}$, respectively, the dual attentions $C_{1}(\cdot)$ and $S_{1}(\cdot)$ are defined as:
\begin{equation}
  C_{1} (\overline{f}) = \delta(Conv(\overline{f}) \times Conv(\overline{f})^{T})^{T}\times Conv(\overline{f}) + \overline{f},
  \label{Equation 3}
\end{equation}
\begin{equation}
  S_{1} (\overline{f}) = Conv(\overline{f}) \times \delta(Conv(\overline{f})^{T} \times Conv(f))^{T} + \overline{f},
  \label{Equation 4}
\end{equation}
where $(\cdot)^{T}$ denotes the transpose operation and $\delta(\cdot)$ is the Softmax function. Note that the channel of the feature in Eq. (\ref{Equation 4}) is set to 1/8 of the original channel of that for computation efficiency.

In the second phase $P_{2}$, for the feature $\widetilde{f}\in \mathbb{R}^{C\times H\times W}$, which can be reshaped with the feature $\overline{f}$, the dual attentions $C_{2}(\cdot)$ and $S_{2}(\cdot)$ are defined as:
\begin{equation}
  C_{2} (\widetilde{f}) = \sigma(MLP(GMP(\widetilde{f}))) \odot \widetilde{f},
  \label{Equation 5}
\end{equation}
\begin{equation}
  S_{2} (\widetilde{f}) = \sigma(Conv(GMPC(\widetilde{f}))) \odot \widetilde{f},
  \label{Equation 6}
\end{equation}
where $\odot$ denotes the element-wise multiplication, $\sigma(\cdot)$ is the Sigmoid function, $MLP(\cdot)$ represents the multi-layer perceptron, $GMP(\cdot)$ and $GMPC(\cdot)$ represent the global max pooling operation and global max pooling along the channel operation, respectively. Note that we use a global max pooling rather than a global average pooling since our goal is to find the area with the biggest visual influence.

\begin{figure*}[!t] \small
  \centering
  \vspace{-0.16cm}
  \subfigure[]{\includegraphics[width=0.24\textwidth]{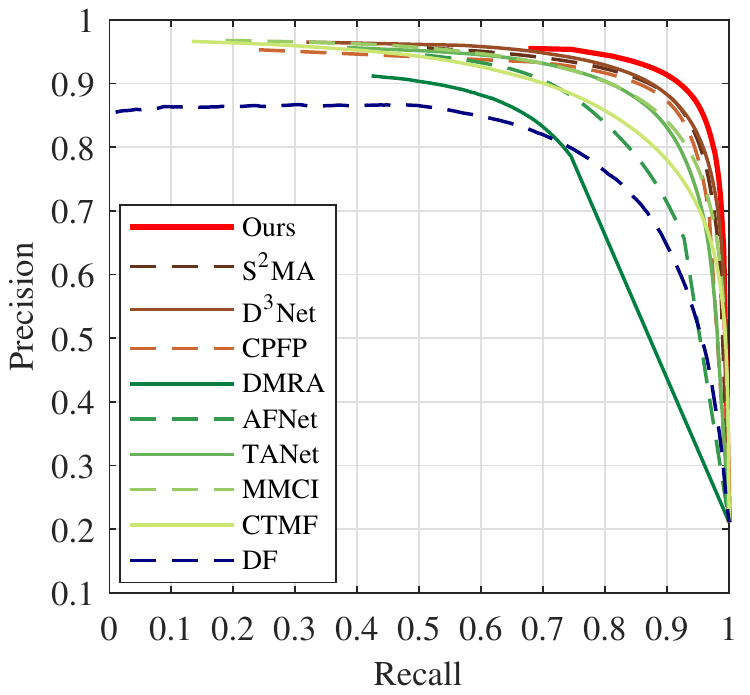}}
  \subfigure[]{\includegraphics[width=0.24\textwidth]{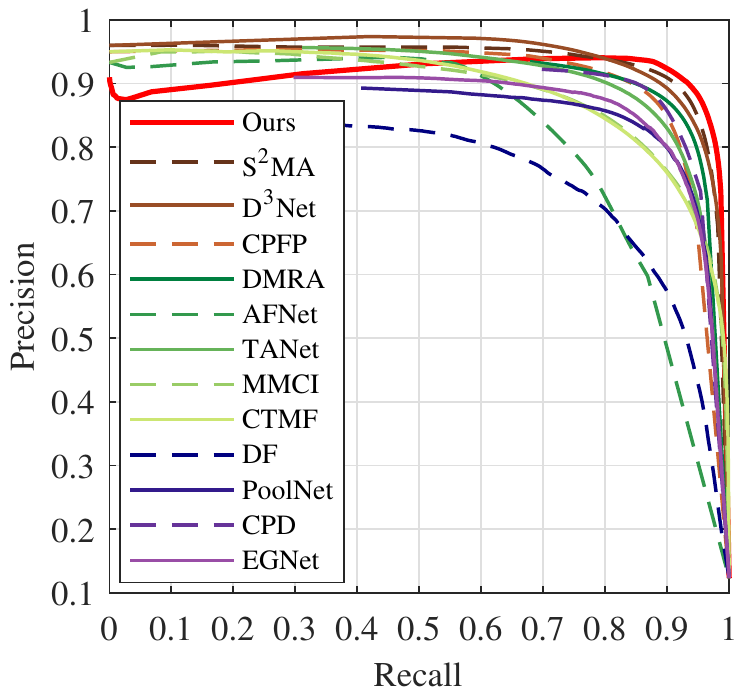}}
  \subfigure[]{\includegraphics[width=0.24\textwidth]{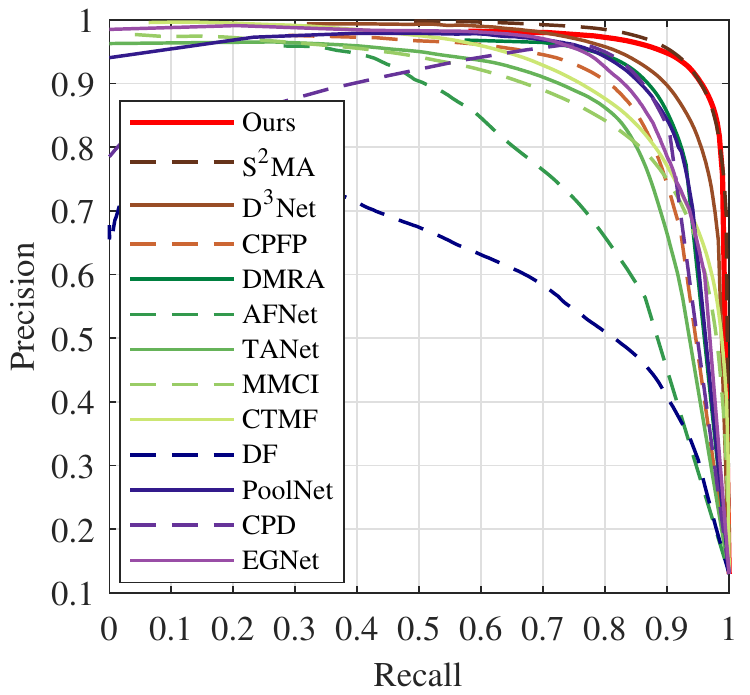}}
  \subfigure[]{\includegraphics[width=0.24\textwidth]{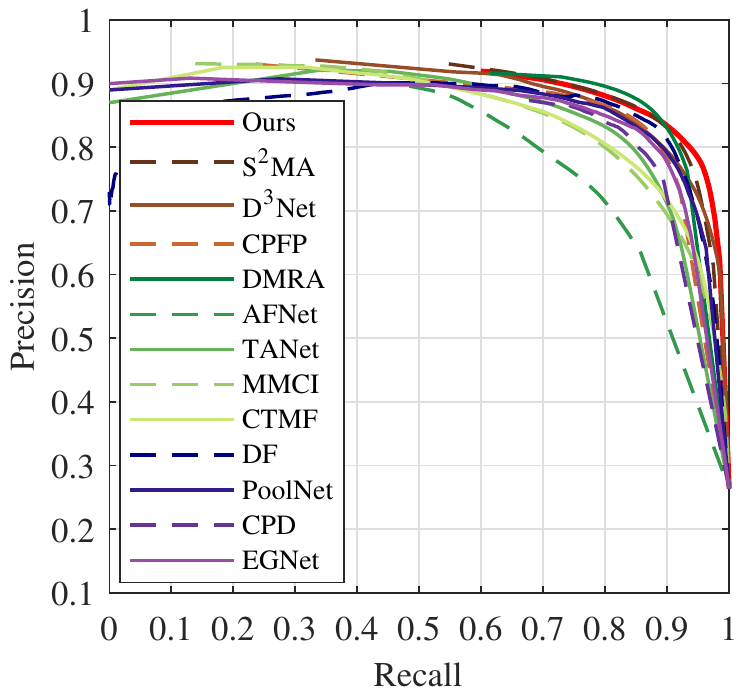}} \\
  \vspace{-0.16cm}
  \subfigure[]{\includegraphics[width=0.24\textwidth]{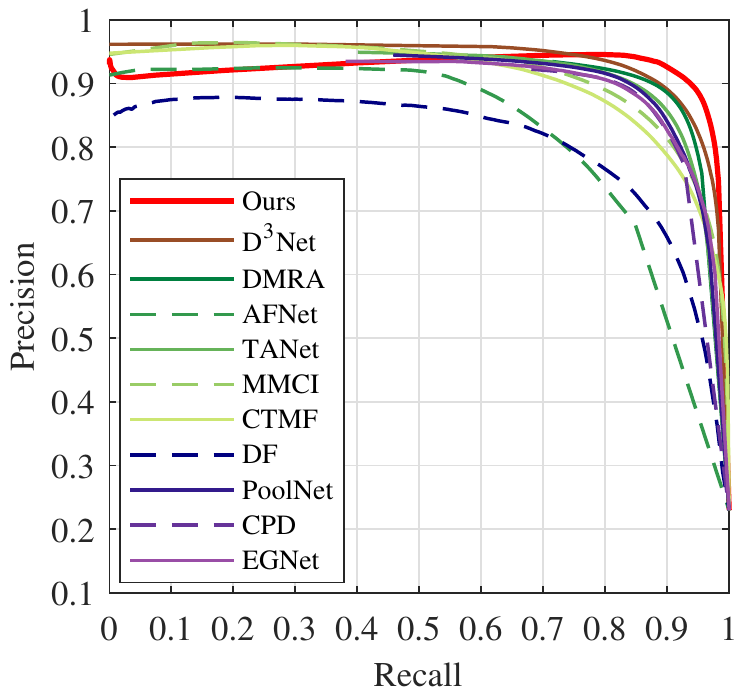}}
  \subfigure[]{\includegraphics[width=0.24\textwidth]{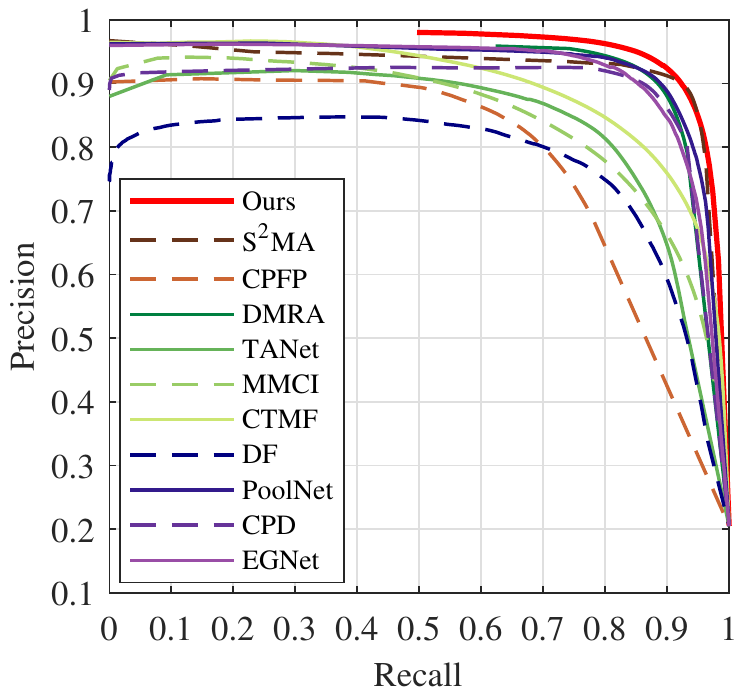}}
  \subfigure[]{\includegraphics[width=0.24\textwidth]{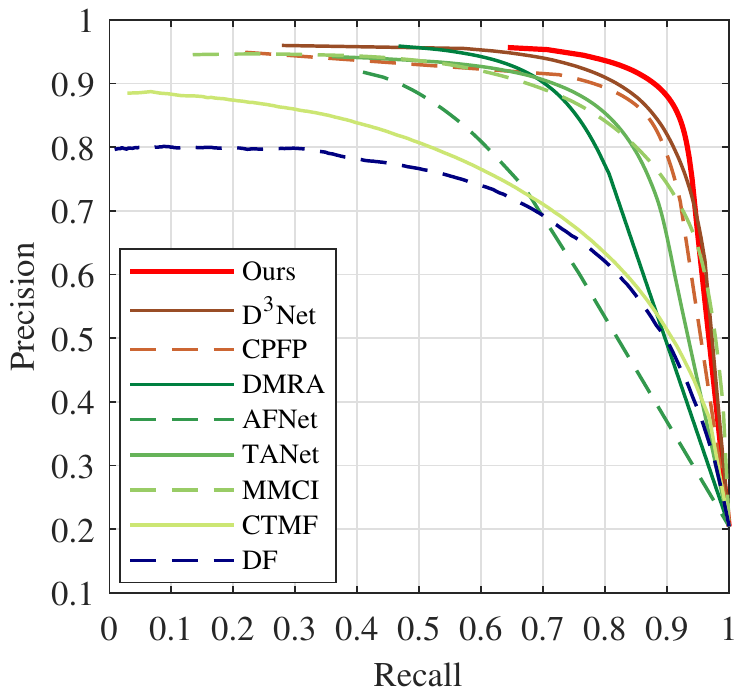}} \\
  \caption{The PR curves of different methods on six datasets. (a) STEREO \cite{Niu2012Leveraging}; (b) NLPR \cite{Peng2014RGBD}; (c) RGBD135 \cite{Cheng2014Depth}; (d) LFSD \cite{Li2014Saliency}; (e) NJU2K \cite{Ju2015Depth-aware}; (f) DUT-RGBD \cite{Piao2019Depth-induced}; (g) SIP\cite{Fan2019Rethinking}.}
  \label{Figure 5}
\end{figure*}

\begin{figure*}[!t] \small
  \centering
  \vspace{-0.16cm}
  \subfigure[]{\includegraphics[width=0.24\textwidth]{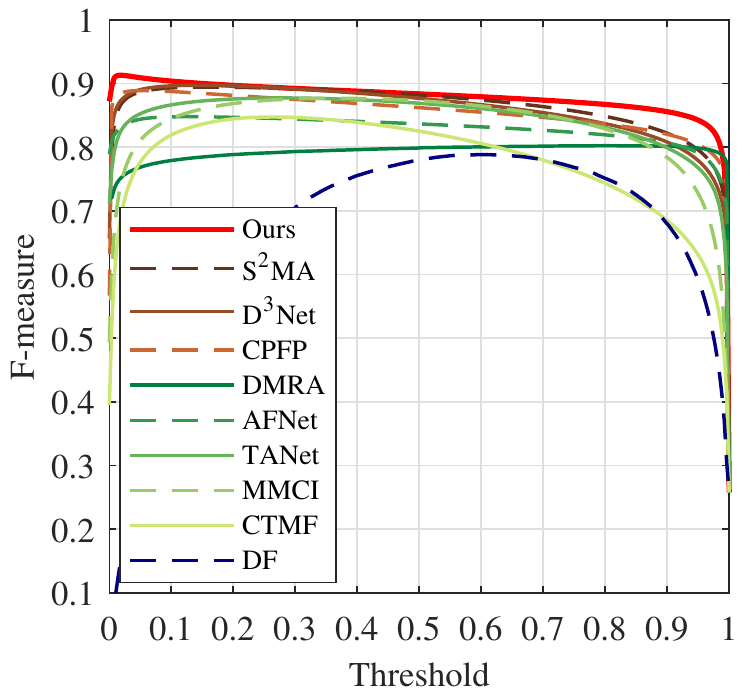}}
  \subfigure[]{\includegraphics[width=0.24\textwidth]{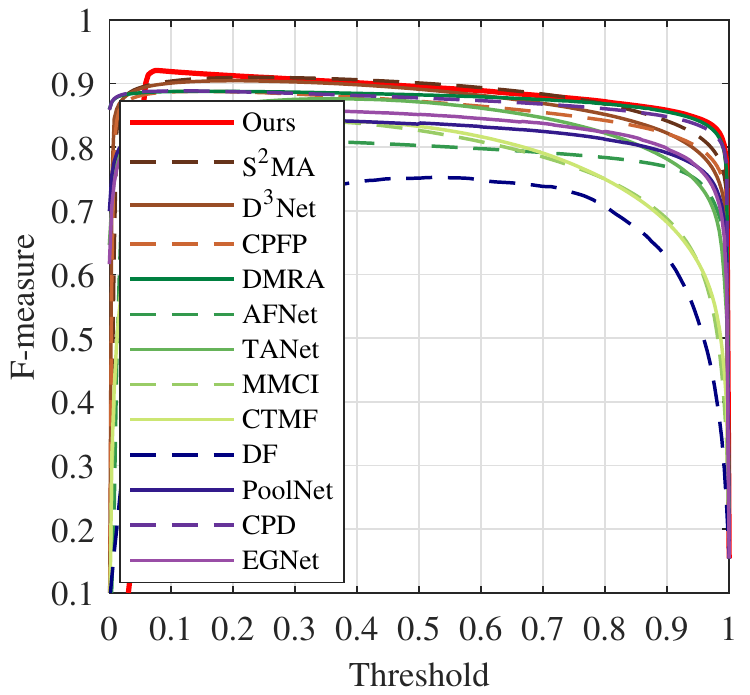}}
  \subfigure[]{\includegraphics[width=0.24\textwidth]{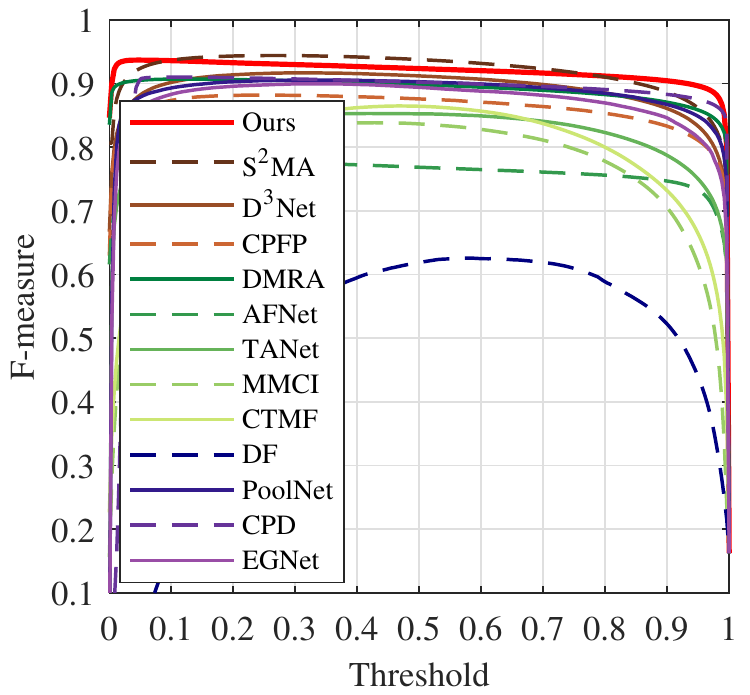}}
  \subfigure[]{\includegraphics[width=0.24\textwidth]{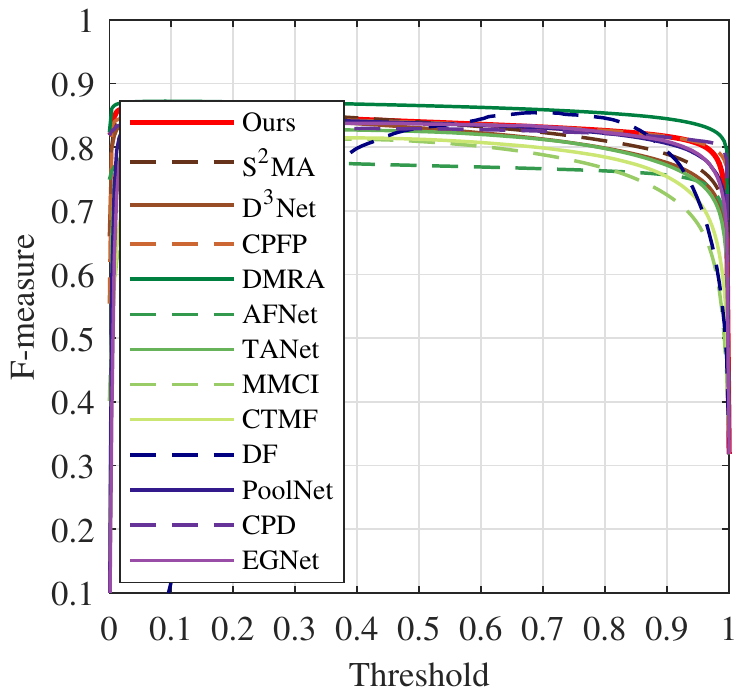}} \\
  \vspace{-0.16cm}
  \subfigure[]{\includegraphics[width=0.24\textwidth]{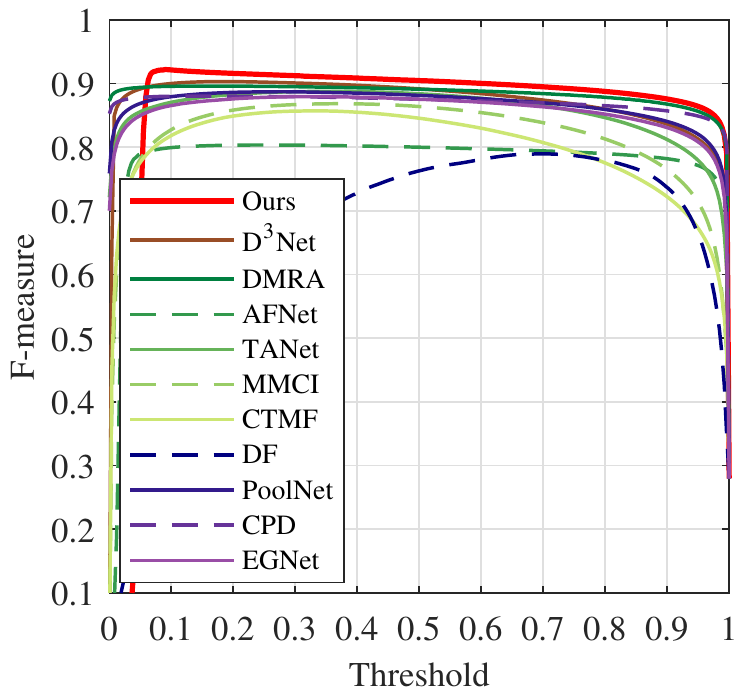}}
  \subfigure[]{\includegraphics[width=0.24\textwidth]{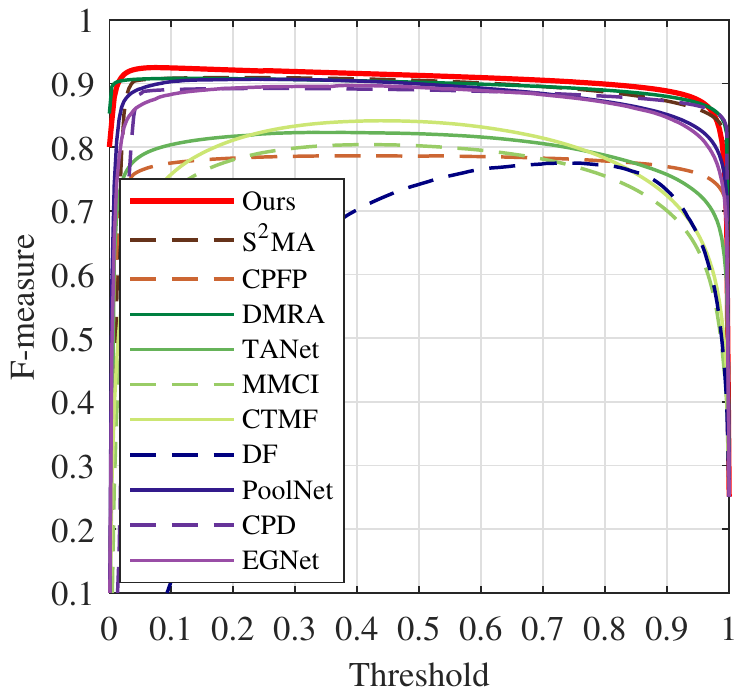}}
  \subfigure[]{\includegraphics[width=0.24\textwidth]{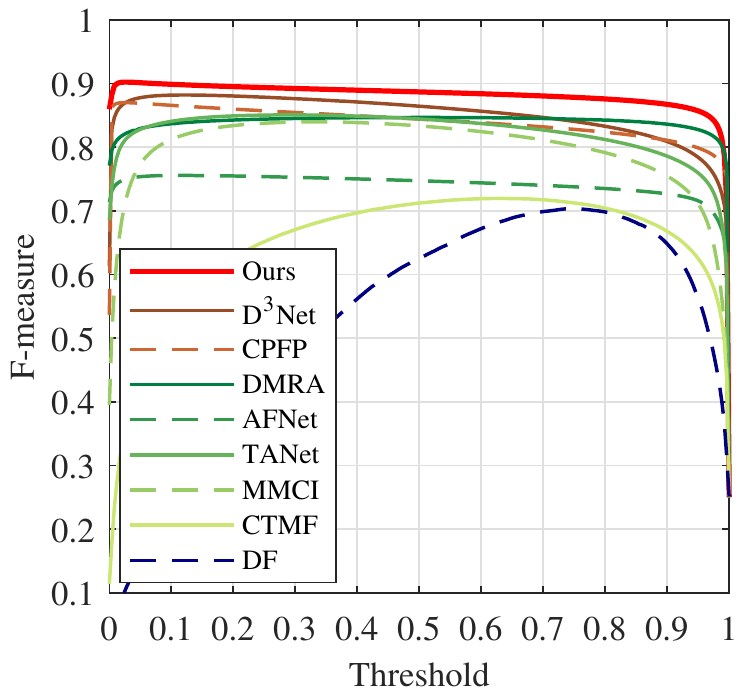}} \\
  \caption{The F$_{\beta}$ curves of different methods on six datasets. (a) STEREO \cite{Niu2012Leveraging}; (b) NLPR \cite{Peng2014RGBD}; (c) RGBD135 \cite{Cheng2014Depth}; (d) LFSD \cite{Li2014Saliency}; (e) NJU2K \cite{Ju2015Depth-aware}; (f) DUT-RGBD \cite{Piao2019Depth-induced}; (g) SIP\cite{Fan2019Rethinking}.}
  \label{Figure 6}
\end{figure*}

\subsection{Adjacent Interactive Aggregation Module}

In general, shallower features have more detail information, while deeper features have more semantic information. The aggregation of multi-level features with different resolutions enables the context information to be integrated as sufficient as possible.

Relying on this, we propose an adjacent interactive aggregation module (AIAM) to guide the interaction of multi-scale features. In this way, the neighbor features that are of intimate correlation are constantly complemented. The interaction behind the features of three triples \{$f_{rgb}^{i-1}$, $f_{rgb}^{i}$, $f_{rgb}^{i+1}$\} $(i = 2, 3, 4)$ helps the generation of desired resulting features \{$f_{rgb^{'}}^{i}$\} $(i = 2, 3, 4)$. The procedure of different levels feature aggregation can be described as:
\begin{equation}
  f_{rgb^{'}}^{i} = I_{1}(f_{rgb}^{i-1}, f_{rgb}^{i}, f_{rgb}^{i+1}) + I_{2}(f_{rgb}^{i-1}, f_{rgb}^{i}, f_{rgb}^{i+1}),
  \label{Equation 7}
\end{equation}
\begin{equation}
  f_{rgbd^{'}}^{i} = Conv(Cat(f_{rgbd}^{i}, f_{rgb^{'}}^{i})),
  \label{Equation 8}
\end{equation}
where $I_{1}$ and $I_{2}$ denote two kinds of feature interactions, as shown in Fig. \ref{Figure 4}.
The difference between them is that the interaction of the former is progressive, while that of the latter is jumping in the initial stage. After that, these relevant features repeatedly pass through some convolutional layers, batch normalization layers and ReLU layers, and finally are associated with a residual block.

\subsection{Joint Hybrid Optimization Loss}

Let $\mathcal{P}=\{p|0<p<1\}$ and $\mathcal{G}=\{g|0<g<1\}$ denote the prediction saliency map and ground truth saliency map, respectively. As the most classical loss function, binary cross entropy loss (BCEL), symbolized by $\mathcal{L}_{BCE}$, can be formulated as:
\begin{equation}
  \mathcal{L}_{BCE} = -\sum\limits_{h=1}^H \sum\limits_{w=1}^W (g\log p + (1-g)\log (1-p)),
  \label{Equation 9}
\end{equation}
where $H$ and $W$ are the height and width of the image, respectively.

\begin{figure*}[ht] \small
  \centering
  \includegraphics[width=0.12\textwidth]{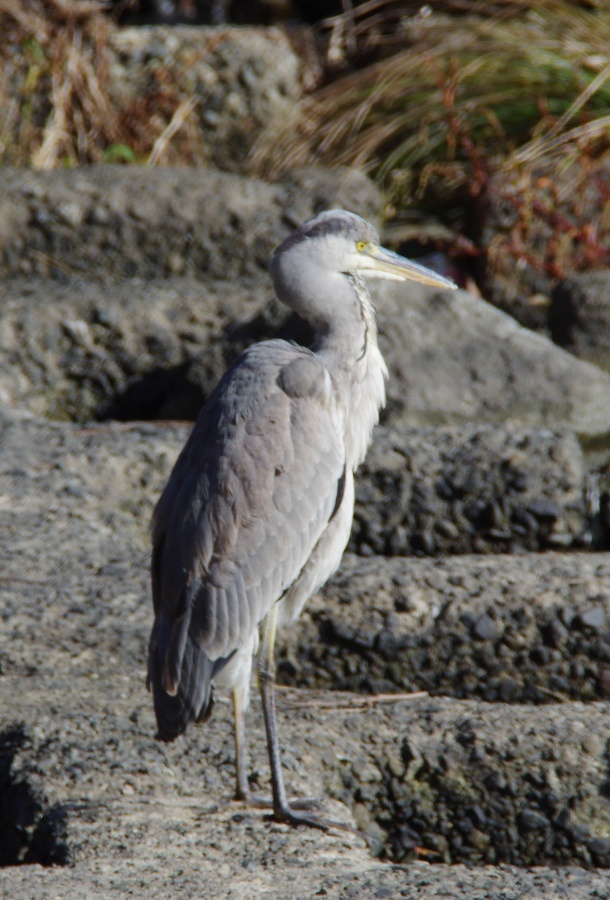}
  \includegraphics[width=0.12\textwidth]{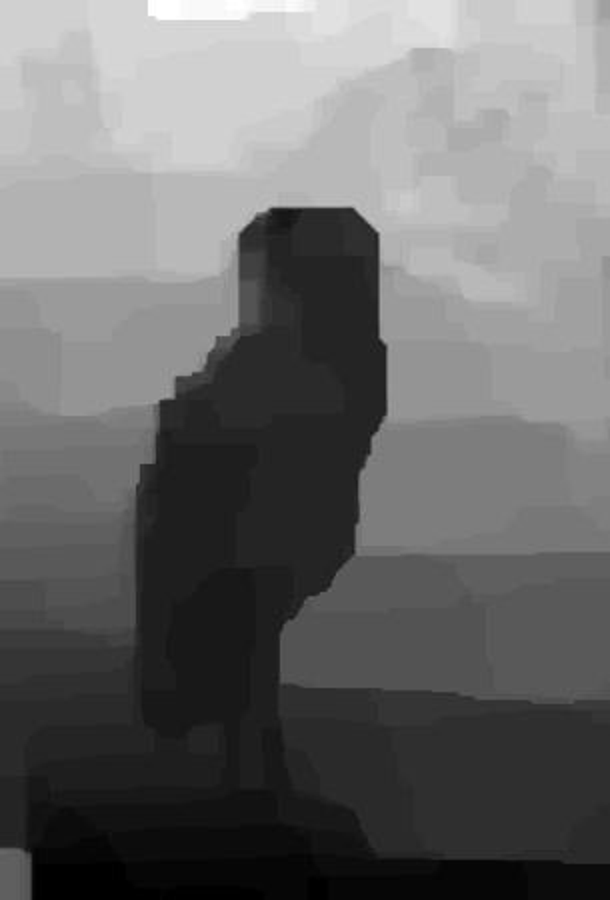}
  \includegraphics[width=0.12\textwidth]{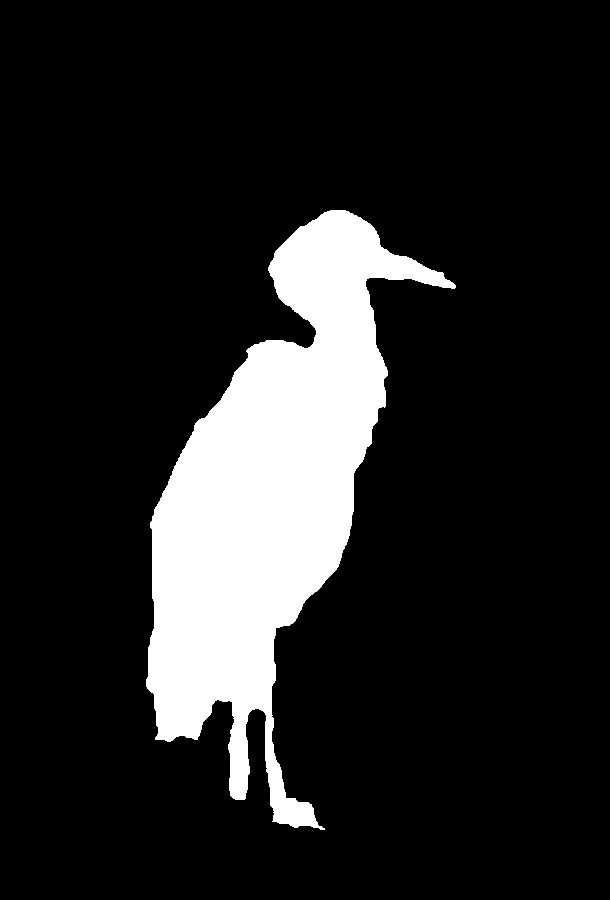}
  \includegraphics[width=0.12\textwidth]{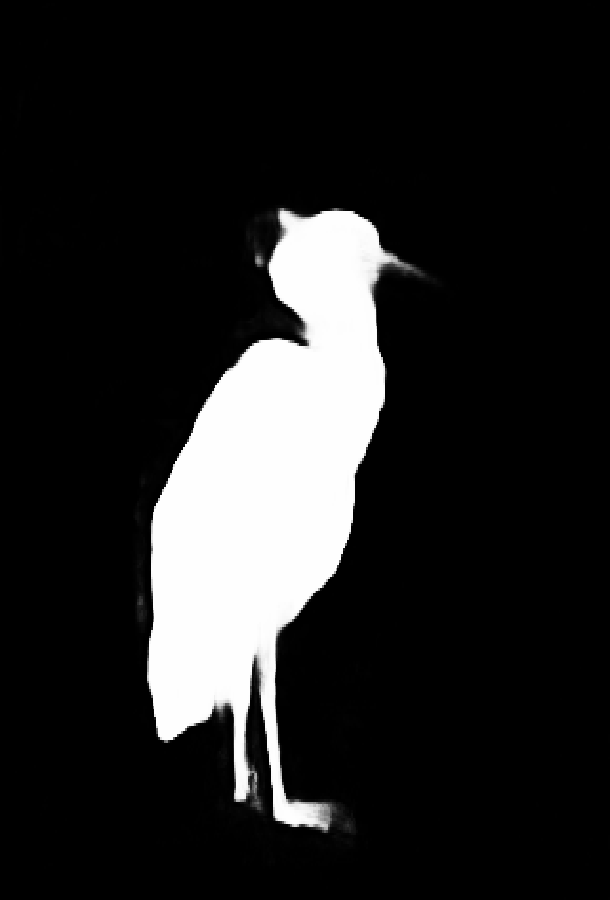}
  \includegraphics[width=0.12\textwidth]{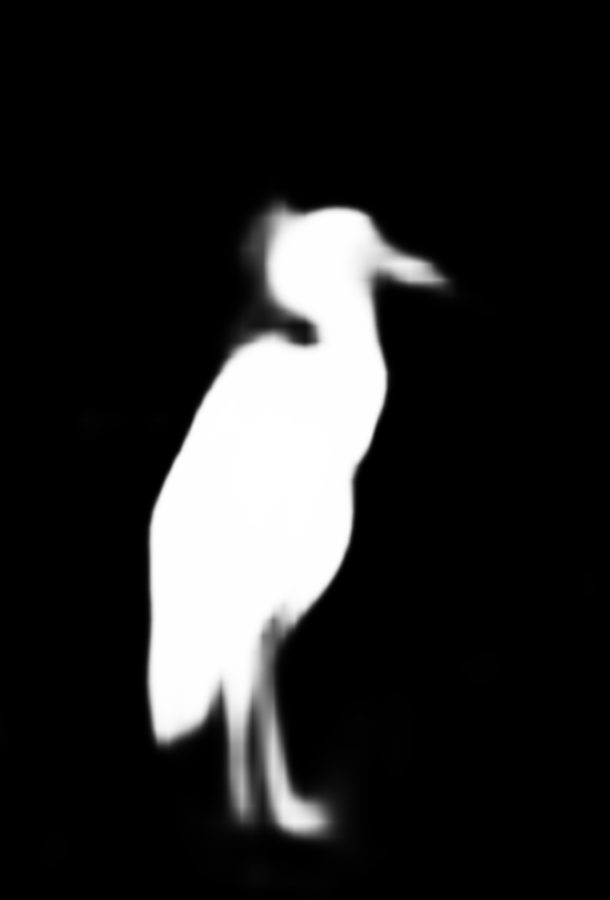}
  \includegraphics[width=0.12\textwidth]{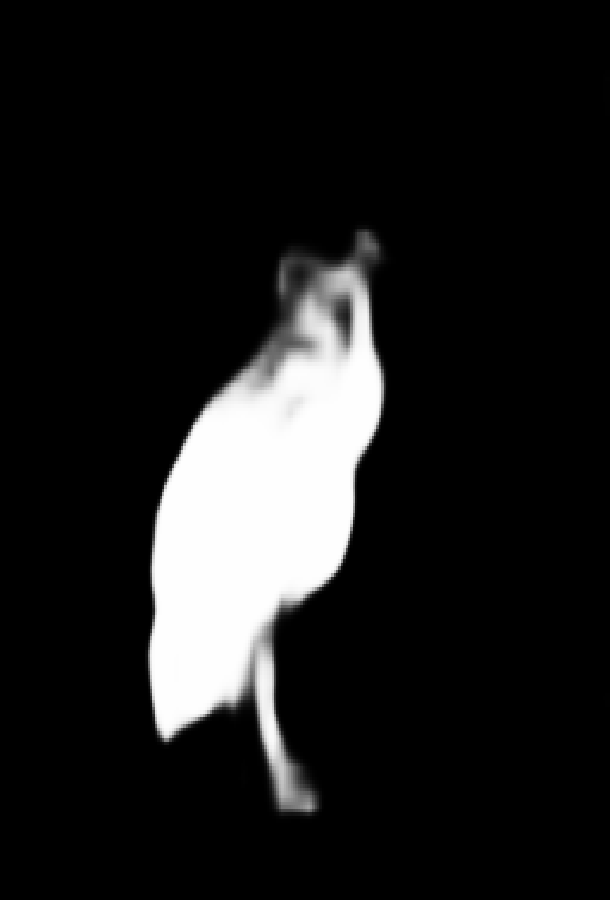}
  \includegraphics[width=0.12\textwidth]{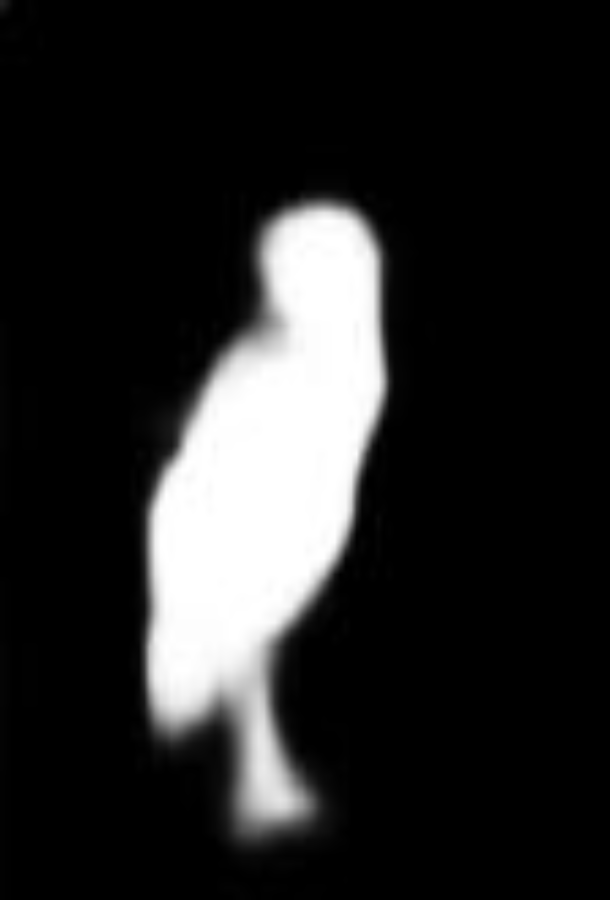}
  \includegraphics[width=0.12\textwidth]{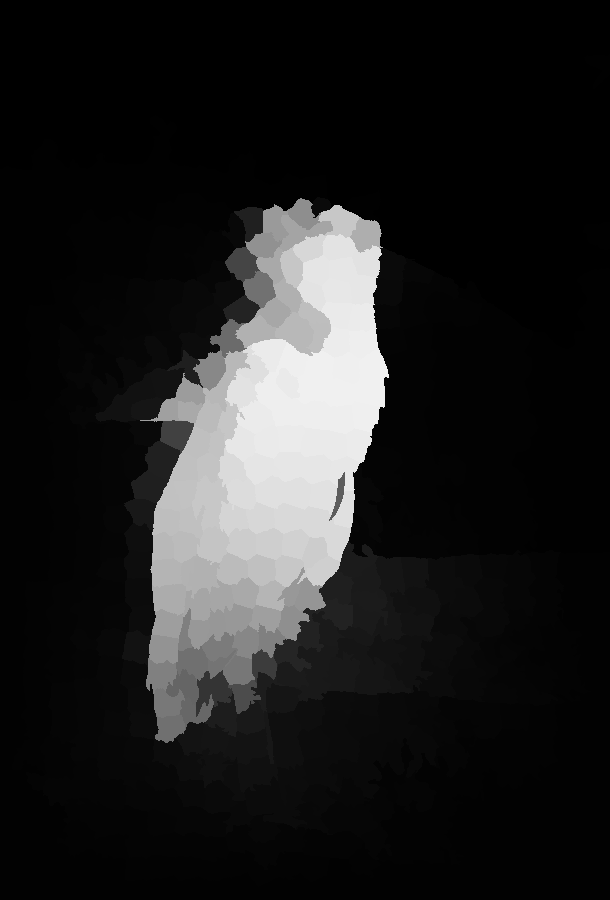} \\
  \vspace{0.025cm}
  \includegraphics[width=0.12\textwidth]{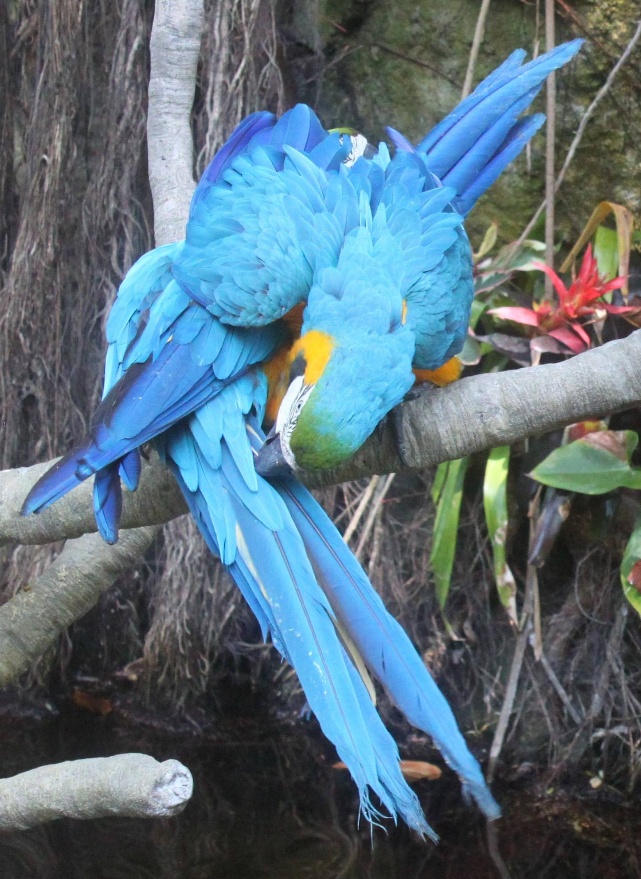}
  \includegraphics[width=0.12\textwidth]{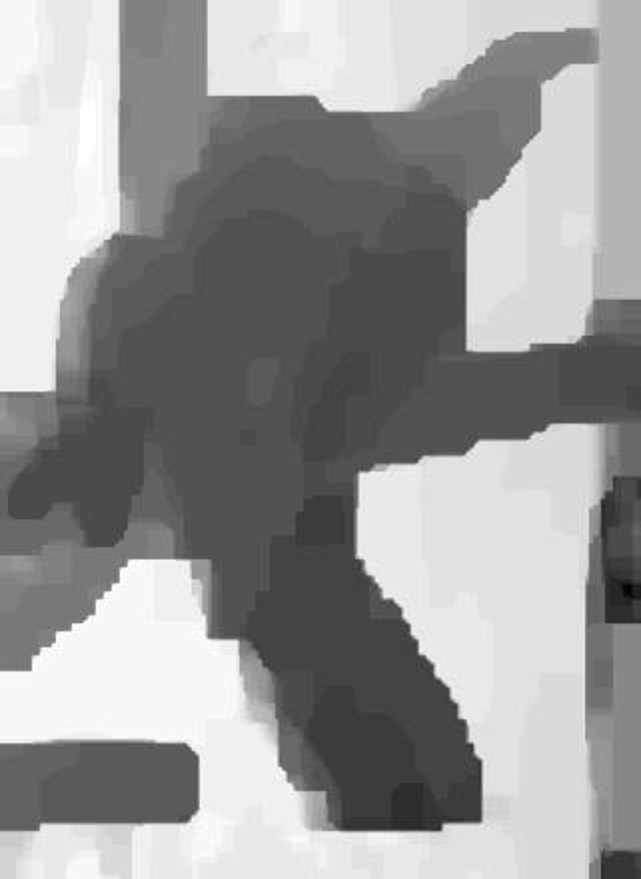}
  \includegraphics[width=0.12\textwidth]{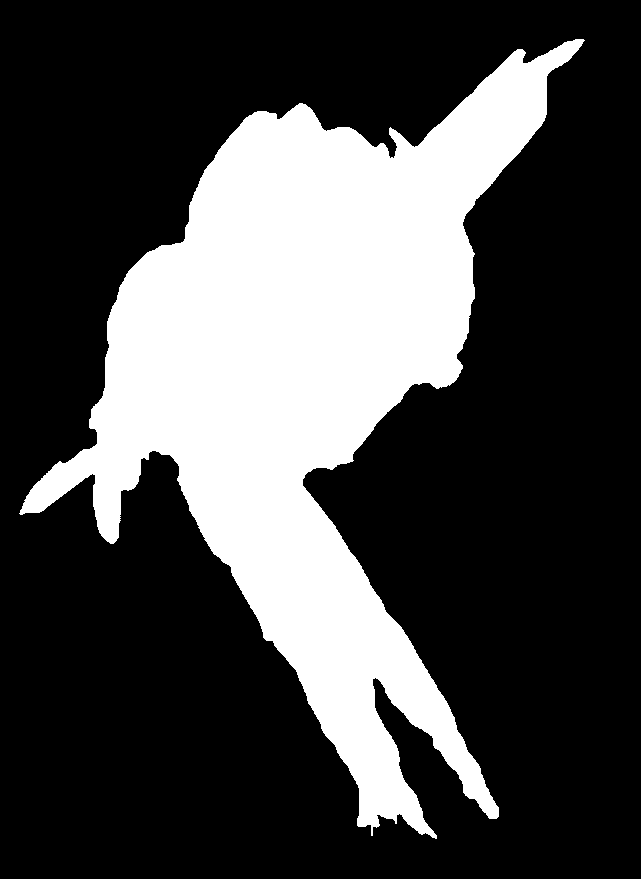}
  \includegraphics[width=0.12\textwidth]{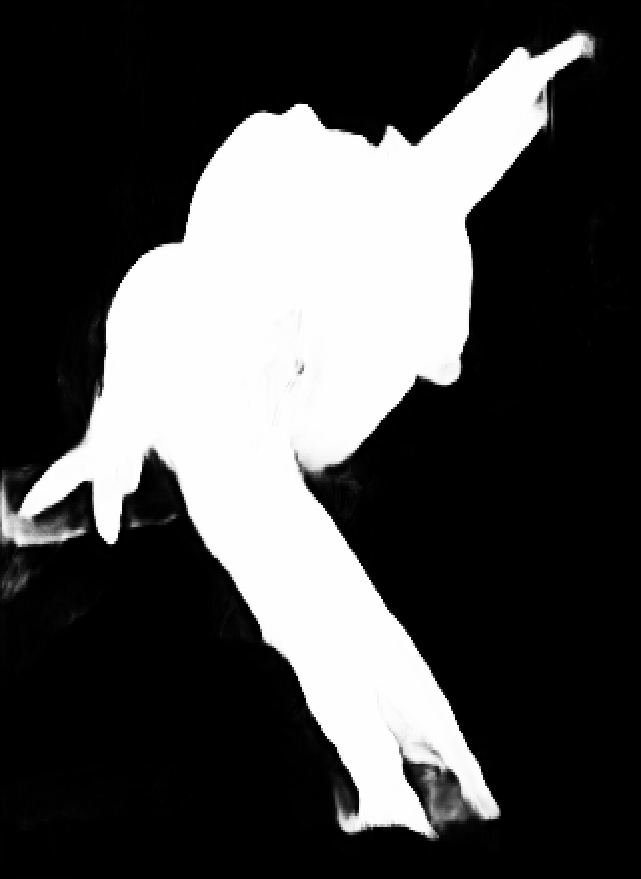}
  \includegraphics[width=0.12\textwidth]{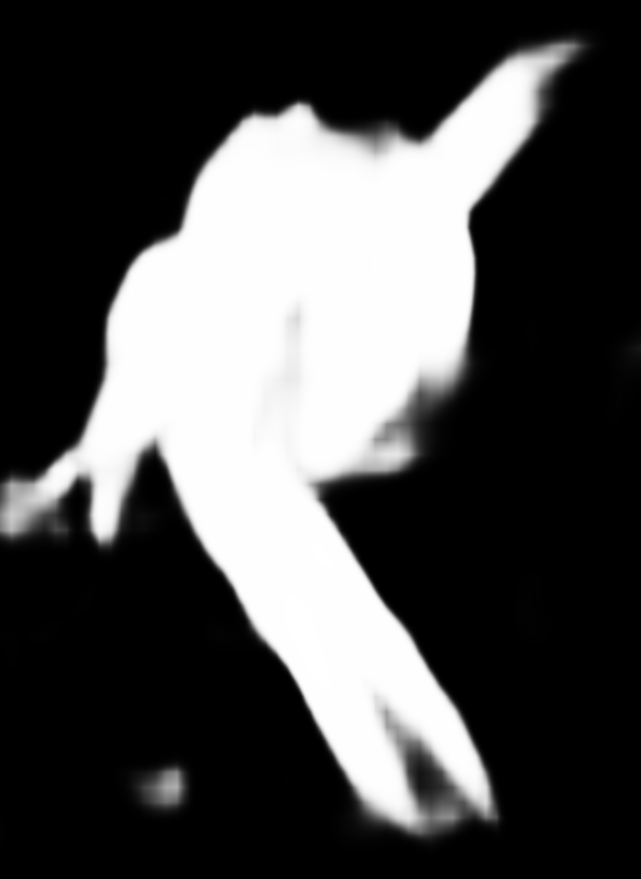}
  \includegraphics[width=0.12\textwidth]{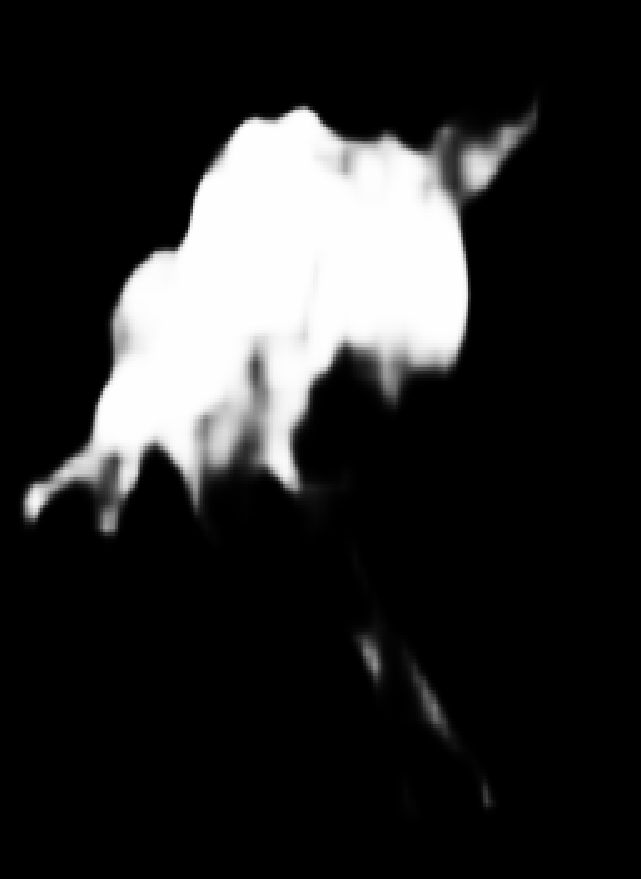}
  \includegraphics[width=0.12\textwidth]{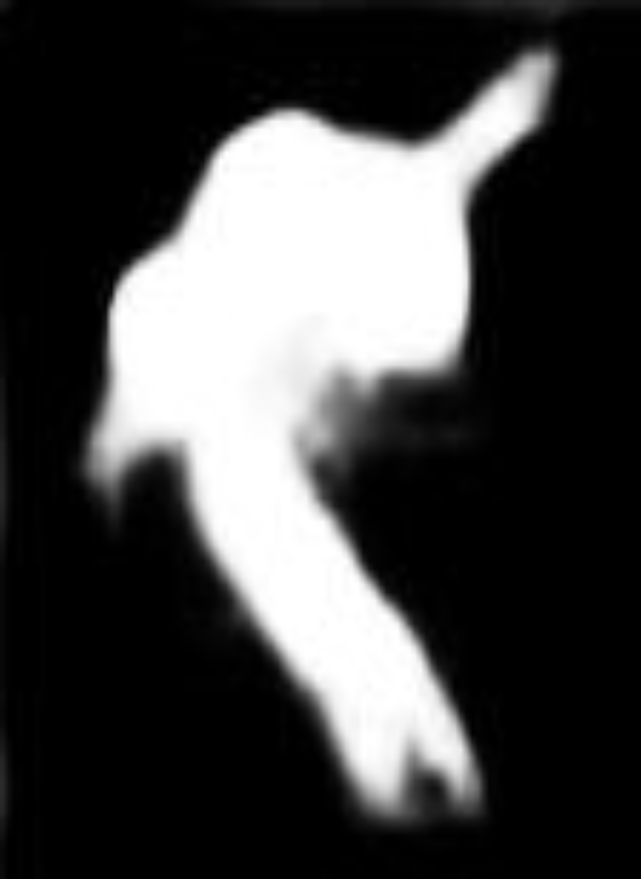}
  \includegraphics[width=0.12\textwidth]{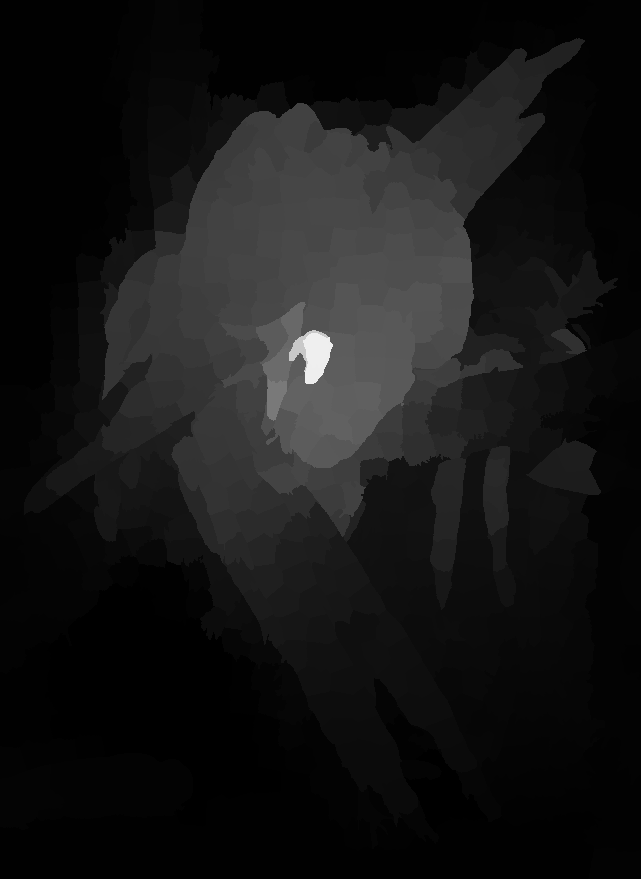} \\
  \vspace{-0.16cm}
  \subfigure[]{\includegraphics[width=0.12\textwidth]{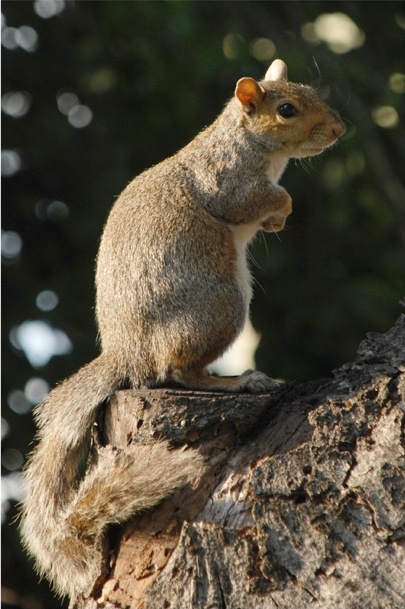}}
  \subfigure[]{\includegraphics[width=0.12\textwidth]{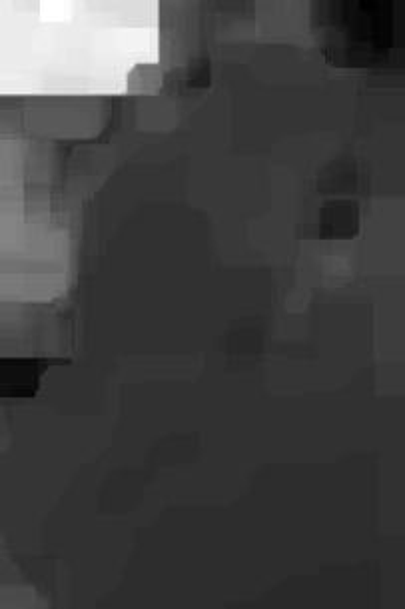}}
  \subfigure[]{\includegraphics[width=0.12\textwidth]{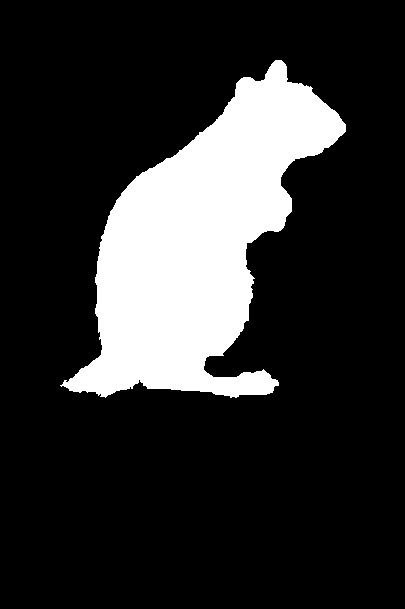}}
  \subfigure[]{\includegraphics[width=0.12\textwidth]{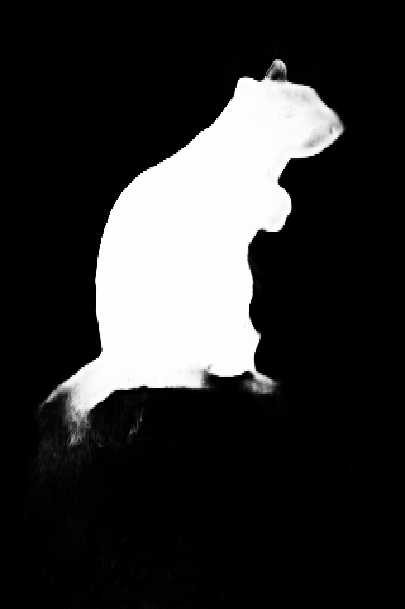}}
  \subfigure[]{\includegraphics[width=0.12\textwidth]{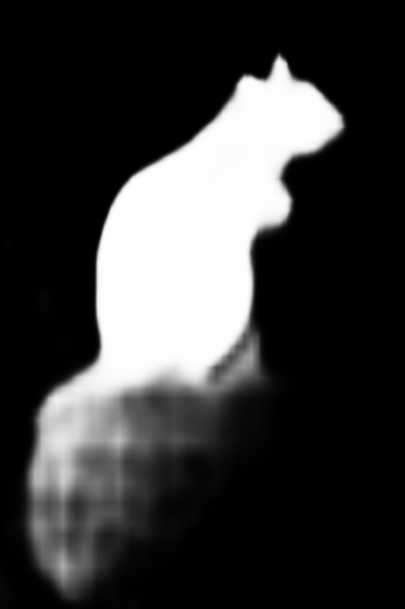}}
  \subfigure[]{\includegraphics[width=0.12\textwidth]{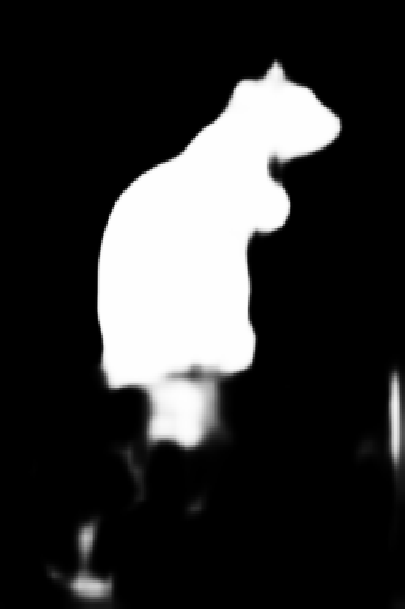}}
  \subfigure[]{\includegraphics[width=0.12\textwidth]{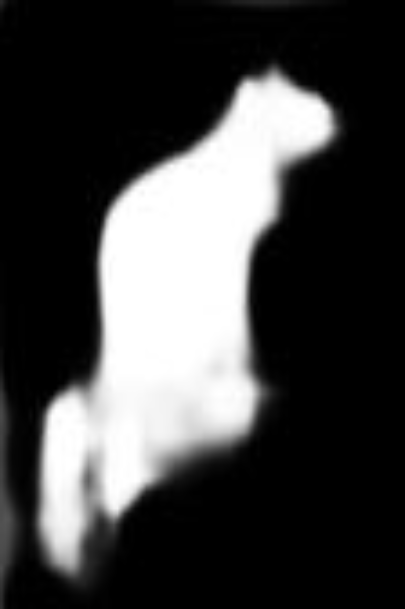}}
  \subfigure[]{\includegraphics[width=0.12\textwidth]{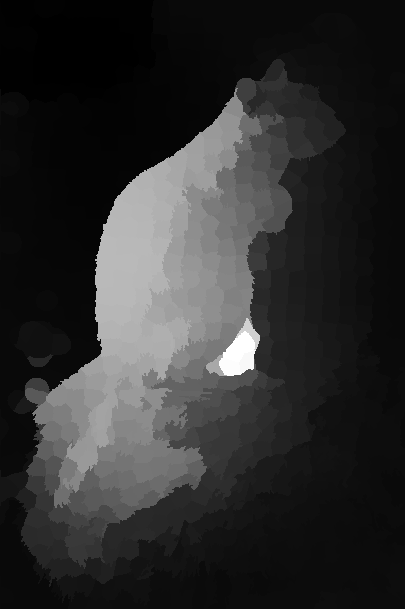}} \\
  \caption{Qualitative comparison of some methods on several representative examples. (a) RGB; (b) Depth; (c) Ground truth; (d) Ours; (e) D$^{3}$Net \cite{Fan2019Rethinking}; (f) DMRA \cite{Piao2019Depth-induced}; (g) TANet \cite{Chen2019Three-stream}; (h) DF \cite{Qu2017RGBD}.}
  \label{Figure 7}
\end{figure*}

To reinforce the capability of supervision from both local and global aspects, we propose a joint hybrid optimization loss (JHOL), symbolized by $\mathcal{L}_{JHO}$, as an auxiliary loss function, which can be formulated as:
\begin{equation}
  \mathcal{L}_{JHO} = \lambda_{1}\mathcal{L}_{1} + \lambda_{2}\mathcal{L}_{2} + \lambda_{3}\mathcal{L}_{3} + \lambda_{4}\mathcal{L}_{4},
  \label{Equation 10}
\end{equation}
where $\lambda_{1}$, $\lambda_{2}$, $\lambda_{3}$ and $\lambda_{4}$ are the parameters to control the trade-off among the four terms of the loss. For the sake of simplicity, they are all set to 1. To be specific, $\mathcal{L}_{1}$, $\mathcal{L}_{2}$, $\mathcal{L}_{3}$ and $\mathcal{L}_{4}$ are defined as:
\begin{equation}
  \mathcal{L}_{1} =  \frac{\sum\limits_{h=1}^H \sum\limits_{w=1}^W (p(1-g))}{\sum\limits_{h=1}^H \sum\limits_{w=1}^W p},
  \label{Equation 11}
\end{equation}
\begin{equation}
  \mathcal{L}_{2} =  \frac{\sum\limits_{h=1}^H \sum\limits_{w=1}^W (g(1-p))}{\sum\limits_{h=1}^H \sum\limits_{w=1}^W g},
  \label{Equation 12}
\end{equation}
\begin{equation}
  \mathcal{L}_{3} =  \frac{\sum\limits_{h=1}^H \sum\limits_{w=1}^W (p(1-g) + g(1-p))}{\sum\limits_{h=1}^H \sum\limits_{w=1}^W (p + g - pg)},
  \label{Equation 13}
\end{equation}
\begin{equation}
  \mathcal{L}_{4} =  \frac{\sum\limits_{h=1}^H \sum\limits_{w=1}^W ((1-p)(1-g))}{\sum\limits_{h=1}^H \sum\limits_{w=1}^W (1 - p(1-g) - g(1-p))}.
  \label{Equation 14}
\end{equation}
On the one hand, $\mathcal{L}_{1}$ and $\mathcal{L}_{2}$ focus on the local correlation of each pixel to ensure a certain degree of discrimination. On the other hand, $\mathcal{L}_{3}$ and $\mathcal{L}_{4}$ focus on the global correlation of each pixel to ensure a certain degree of consistency. As a result, $\mathcal{L}_{JHO}$ jointly makes the model capture the foreground region as smoothly as possible and filter the background region as steadily as possible.

Therefore, the total loss $\mathcal{L}_{Total}$ can be written as:
\begin{equation}
  \mathcal{L}_{Total} = \mathcal{L}_{BCE} + \mu\mathcal{L}_{JHO},
  \label{Equation 15}
\end{equation}
where $\mu$ is the parameter to control the trade-off between $\mathcal{L}_{BCE}$ and $\mathcal{L}_{JHO}$. In practice, it is also set to 1.

\begin{table*}[ht] \small
  \caption{Ablation study of our method in terms of uniform evaluation metrics. $\uparrow$ and $\downarrow$ indicate that the larger and smaller scores are better, respectively. The best results are \bf{bold}.}
  \label{Table 2}
  \centering
  \vspace{0.1cm}
  \resizebox{\textwidth}{!}{
    \begin{tabular}{c|c|cc|cc|cccc|cccccc}
      \hline
      \multirow{2}{*}{No.} &\multirow{2}{*}{Baseline}
      &\multicolumn{2}{c|}{+NDAM} &\multicolumn{2}{c|}{+AIAM} &\multicolumn{4}{c|}{+JHOL}
      &\multirow{2}{*}{$S_{\alpha} \uparrow$} &\multirow{2}{*}{$F^{max}_{\beta} \uparrow$} &\multirow{2}{*}{$F^{avg}_{\beta} \uparrow$} &\multirow{2}{*}{$F^{\omega}_{\beta}$} &\multirow{2}{*}{$E_{\xi} \uparrow$} &\multirow{2}{*}{$\mathcal{M} \downarrow$} \\
      \cline{3-10}
      & &+$P_{1}$ &+$P_{2}$ &+$I_{1}$ &+$I_{2}$ &+$\mathcal{L}_{1}$ &+$\mathcal{L}_{2}$ &+$\mathcal{L}_{3}$ &+$\mathcal{L}_{4}$ \\
      \hline
      1  &$\checkmark$ & & & & & & & & &0.885 &0.820 &0.775 &0.899 &0.872 &0.064 \\
      2  &$\checkmark$ &$\checkmark$ & & & & & & & &0.893 &0.833 &0.790 &0.907 &0.880 &0.060 \\
      3  &$\checkmark$ & &$\checkmark$ & & & & & & &0.887 &0.827 &0.779 &0.903 &0.875 &0.063 \\
      4  &$\checkmark$ &$\checkmark$ &$\checkmark$ & & & & & & &0.894 &0.837 &0.796 &0.910 &0.882 &0.059 \\
      5  &$\checkmark$ & & &$\checkmark$ & & & & & &0.896 &0.839 &0.804 &0.914 &0.885 &0.058 \\
      6  &$\checkmark$ & & & &$\checkmark$ & & & & &0.890 &0.832 &0.788 &0.906 &0.878 &0.061 \\
      7  &$\checkmark$ & & &$\checkmark$ &$\checkmark$ & & & & &0.891 &0.865 &0.812 &0.919 &0.873 &0.055 \\
      8  &$\checkmark$ & & & & &$\checkmark$ & & & &0.891 &0.865 &0.812 &0.919 &0.873 &0.055 \\
      9  &$\checkmark$ & & & & & &$\checkmark$ & & &0.892 &0.812 &0.786 &0.899 &0.866 &0.061 \\
      10 &$\checkmark$ & & & & & & &$\checkmark$ & &0.893 &0.848 &0.819 &0.916 &0.876 &0.053 \\
      11 &$\checkmark$ & & & & & & & &$\checkmark$ &0.887 &0.819 &0.778 &0.900 &0.871 &0.063 \\
      12 &$\checkmark$ & & & & &$\checkmark$ &$\checkmark$ &$\checkmark$ &$\checkmark$ &0.899 &0.856 &0.831 &0.921 &0.880 &0.050 \\
      13 &$\checkmark$ &$\checkmark$ &$\checkmark$ &$\checkmark$ &$\checkmark$ &$\checkmark$ &$\checkmark$ &$\checkmark$ &$\checkmark$ &\bf{0.913} &\bf{0.873} &\bf{0.854} &\bf{0.929} &\bf{0.897} &\bf{0.043} \\
      \hline
    \end{tabular}}
  \label{bigtable}
\end{table*}

\begin{figure*} \small
  \centering
  \includegraphics[width=0.12\textwidth]{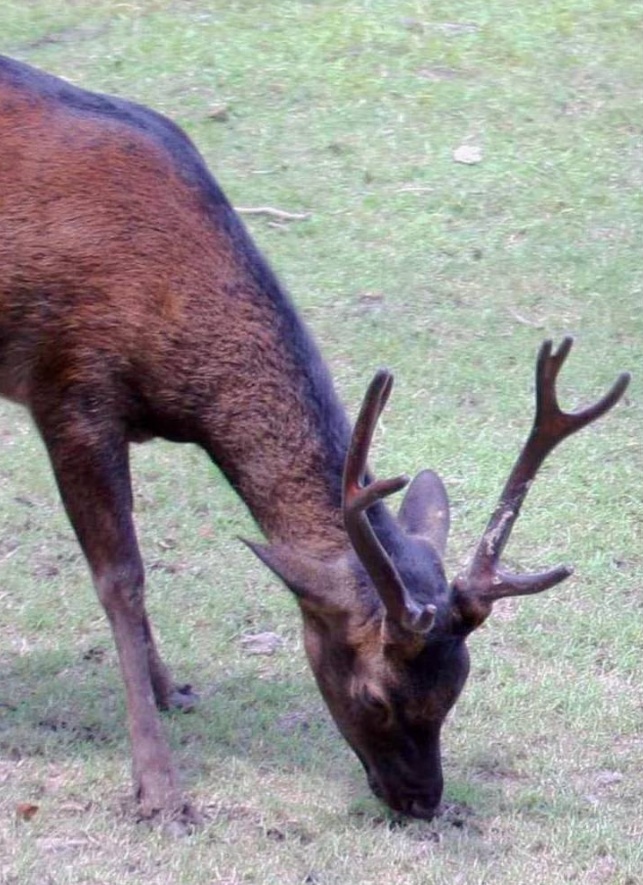}
  \includegraphics[width=0.12\textwidth]{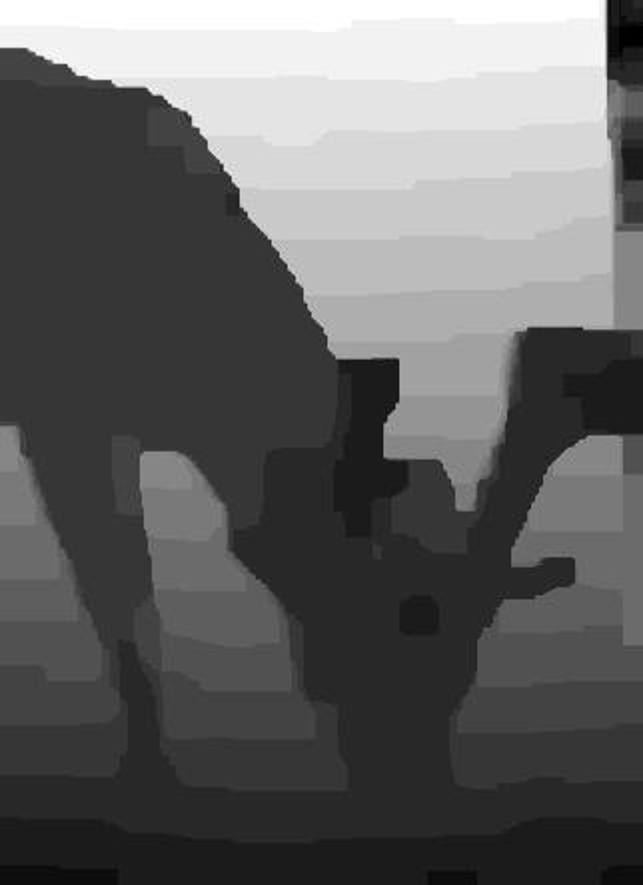}
  \includegraphics[width=0.12\textwidth]{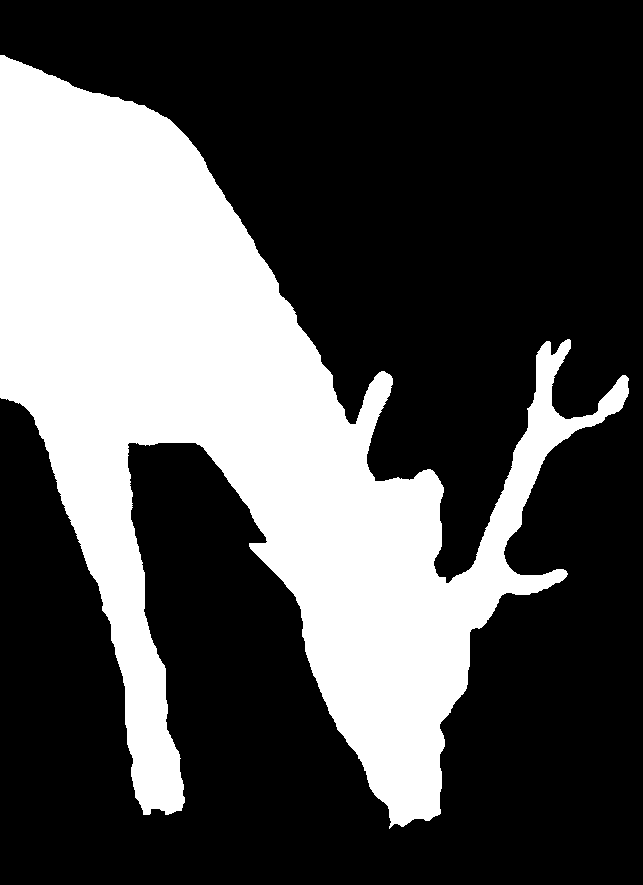}
  \includegraphics[width=0.12\textwidth]{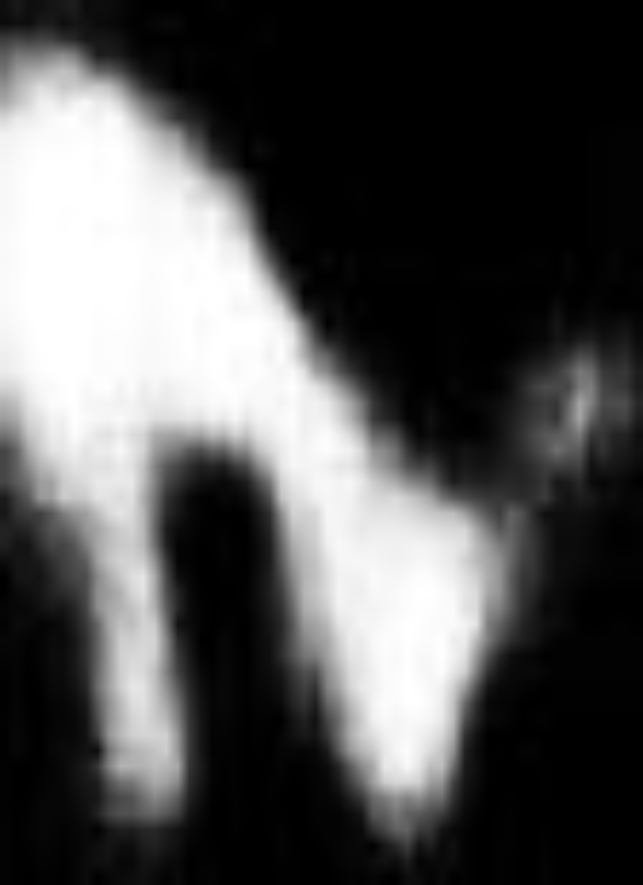}
  \includegraphics[width=0.12\textwidth]{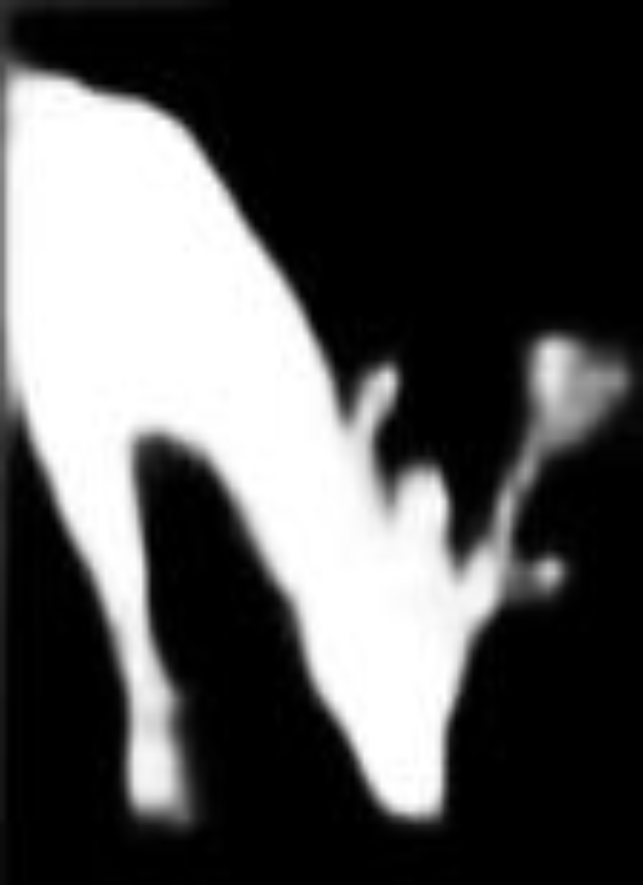}
  \includegraphics[width=0.12\textwidth]{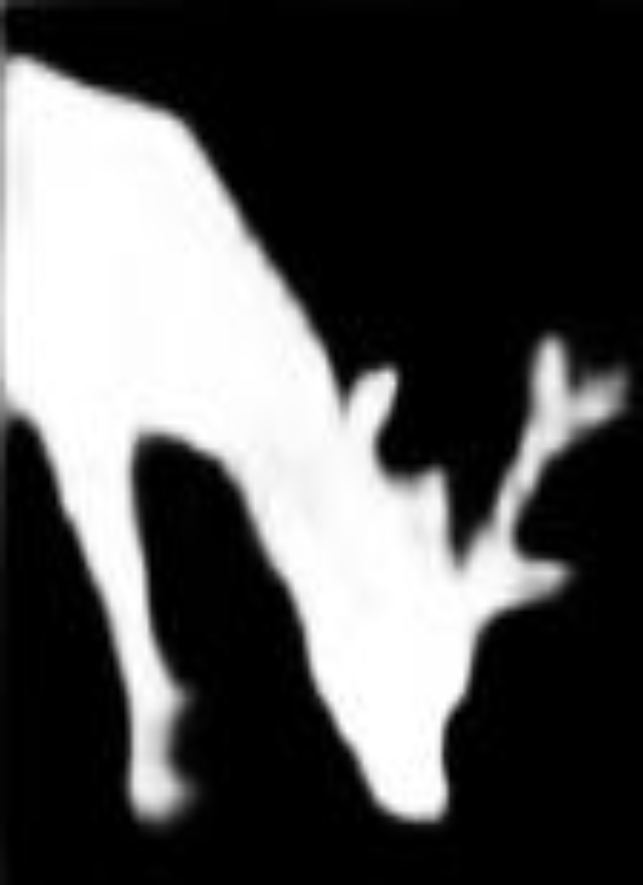}
  \includegraphics[width=0.12\textwidth]{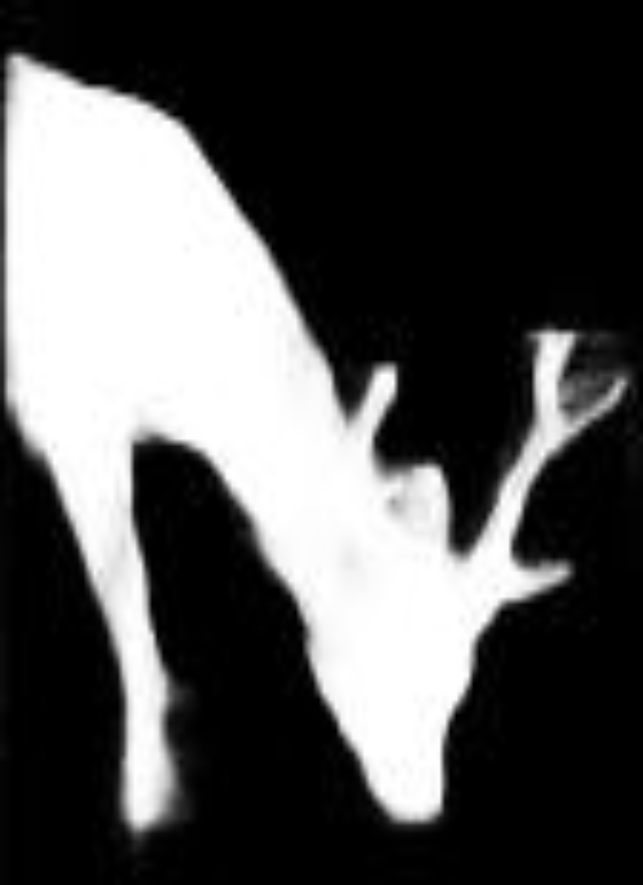}
  \includegraphics[width=0.12\textwidth]{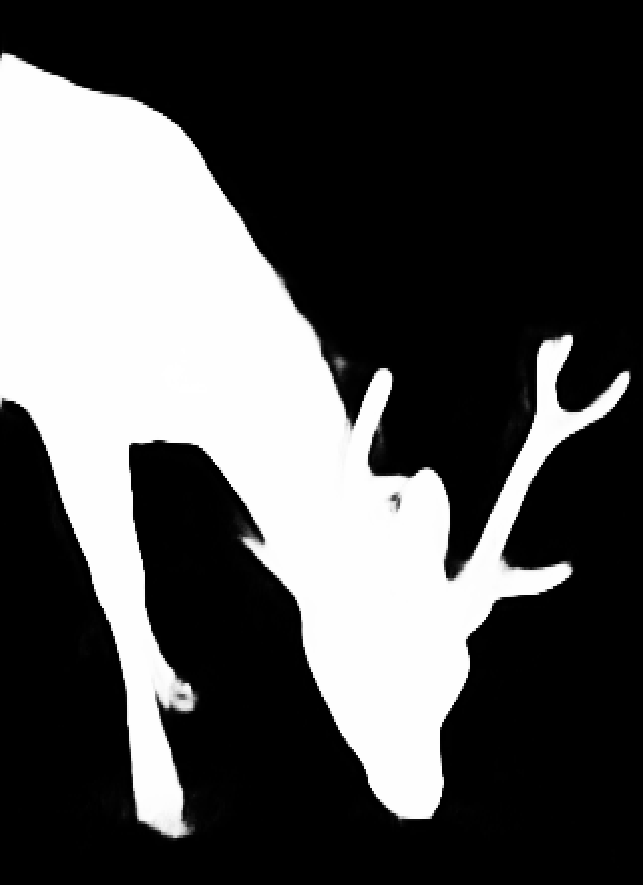} \\
  \vspace{-0.16cm}
  \subfigure[]{\includegraphics[width=0.12\textwidth]{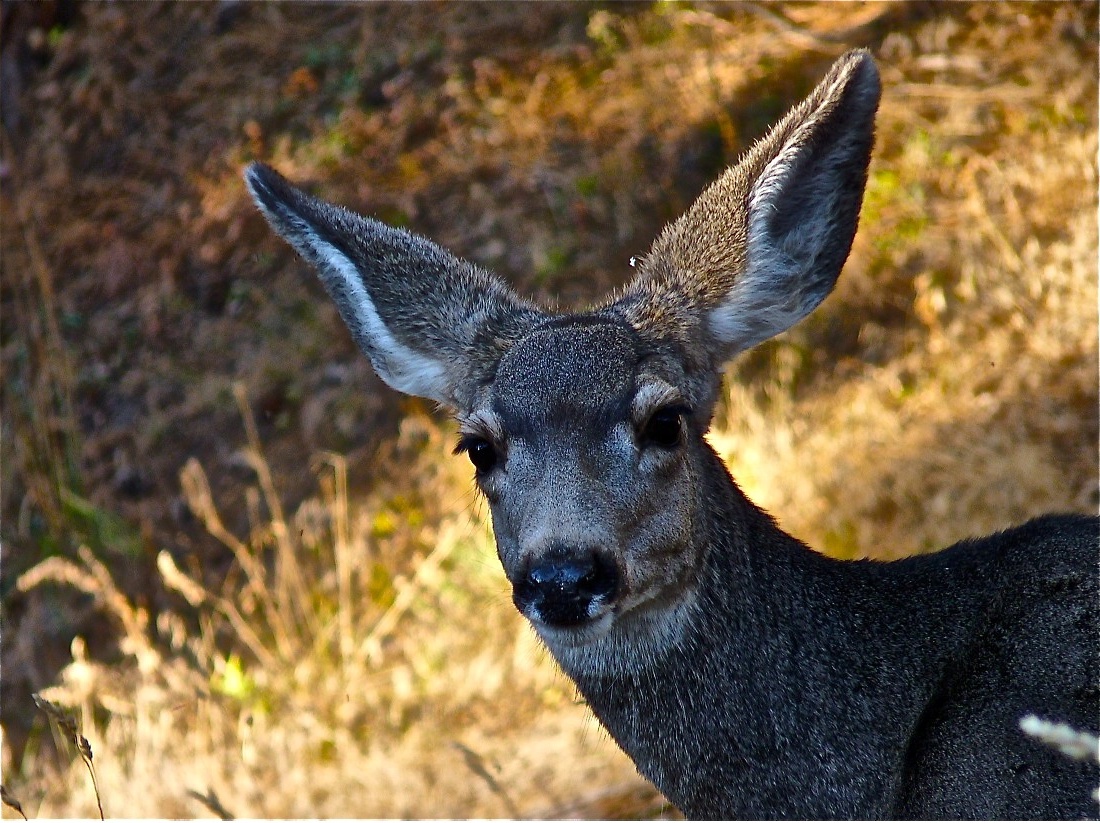}}
  \subfigure[]{\includegraphics[width=0.12\textwidth]{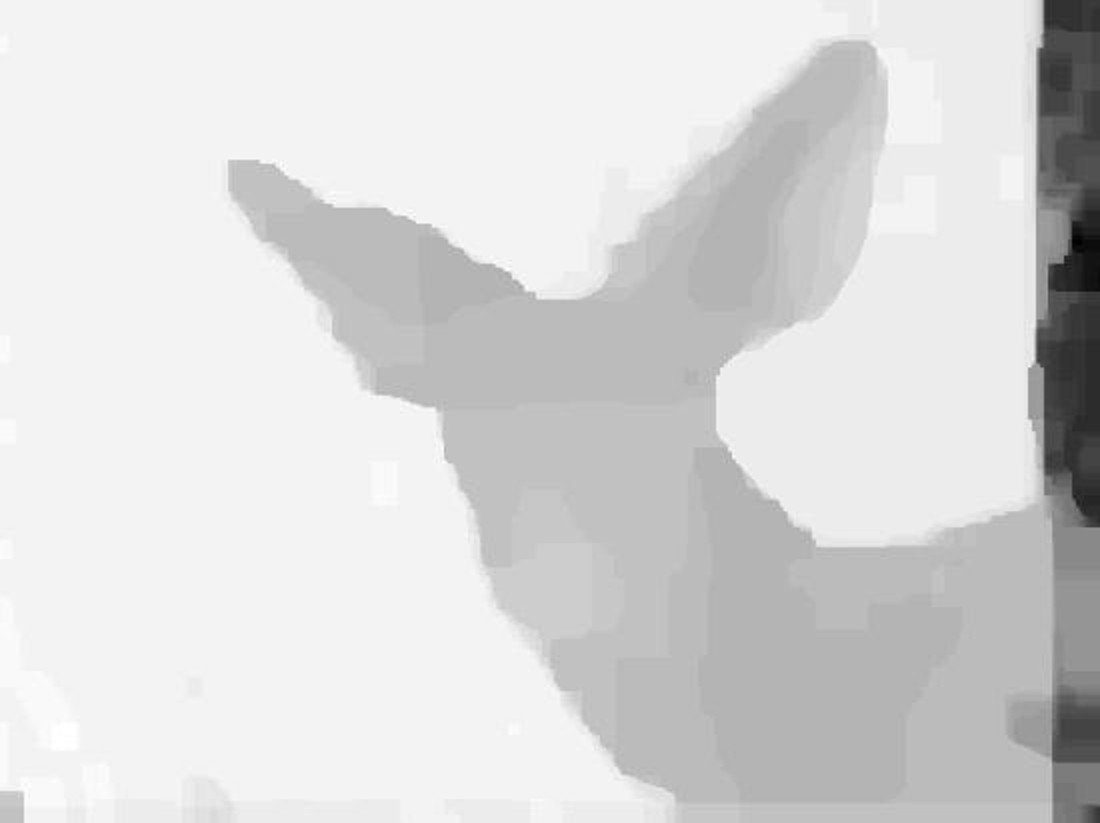}}
  \subfigure[]{\includegraphics[width=0.12\textwidth]{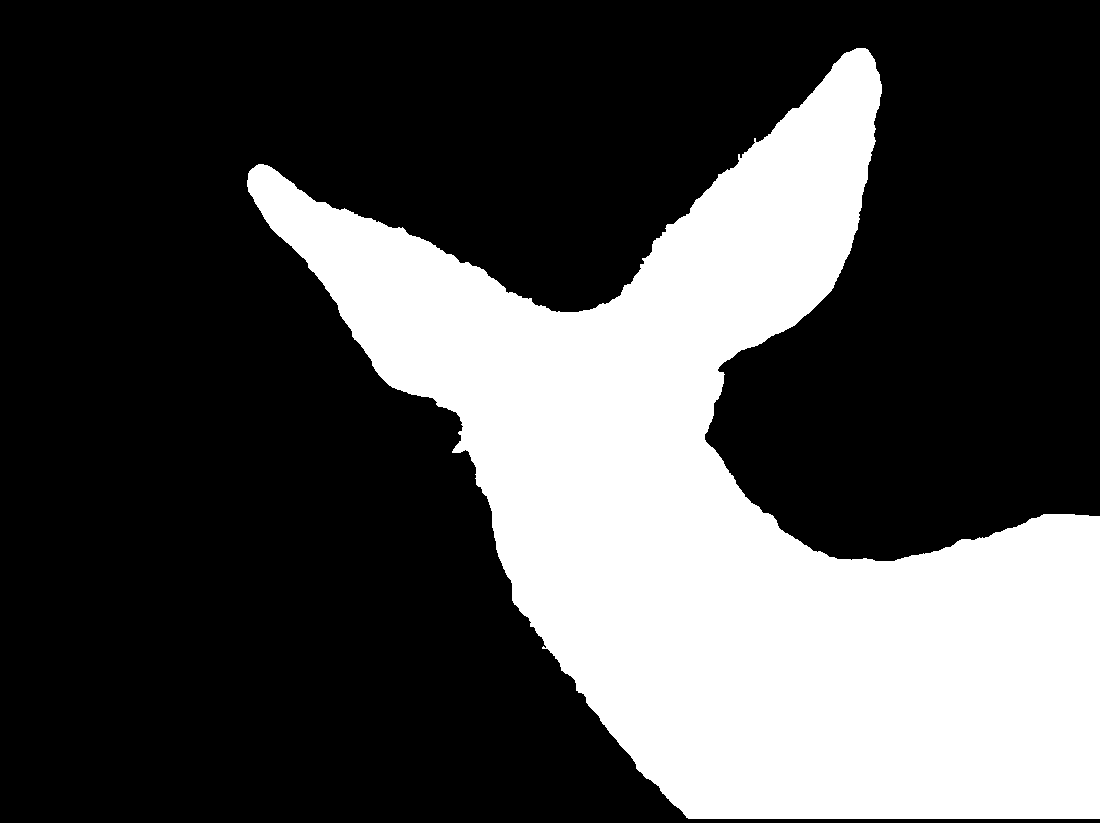}}
  \subfigure[]{\includegraphics[width=0.12\textwidth]{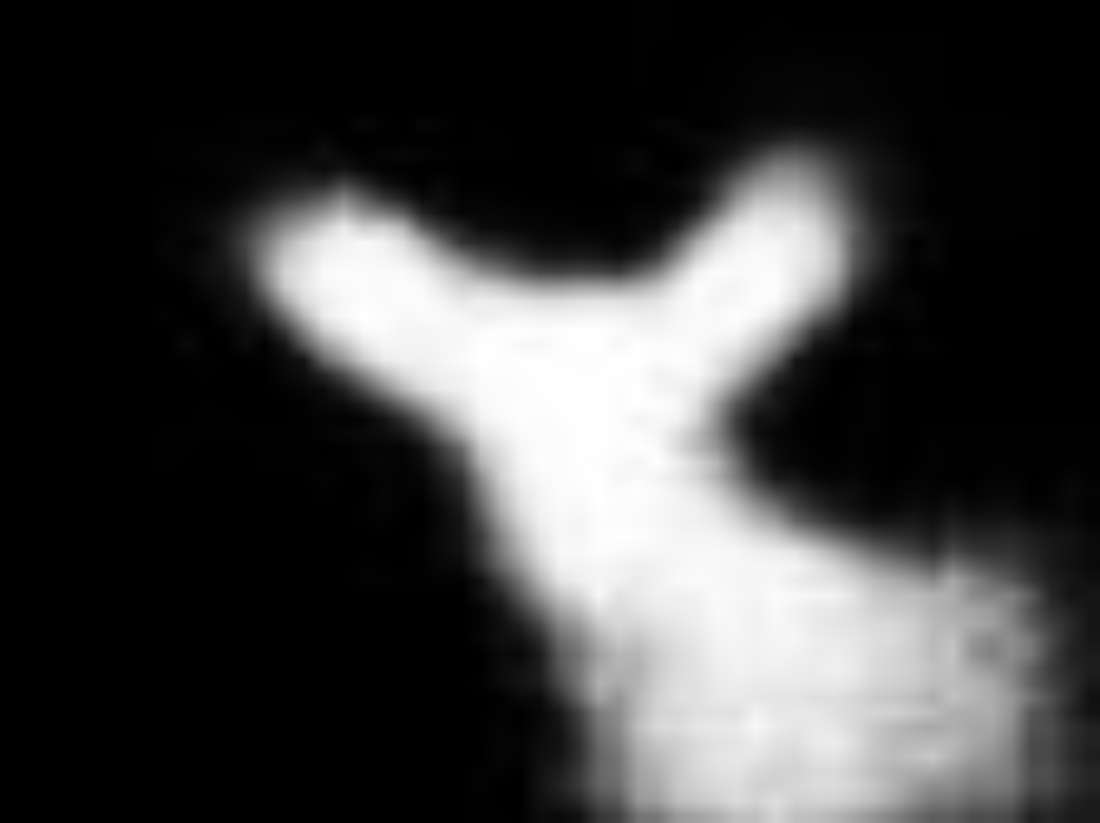}}
  \subfigure[]{\includegraphics[width=0.12\textwidth]{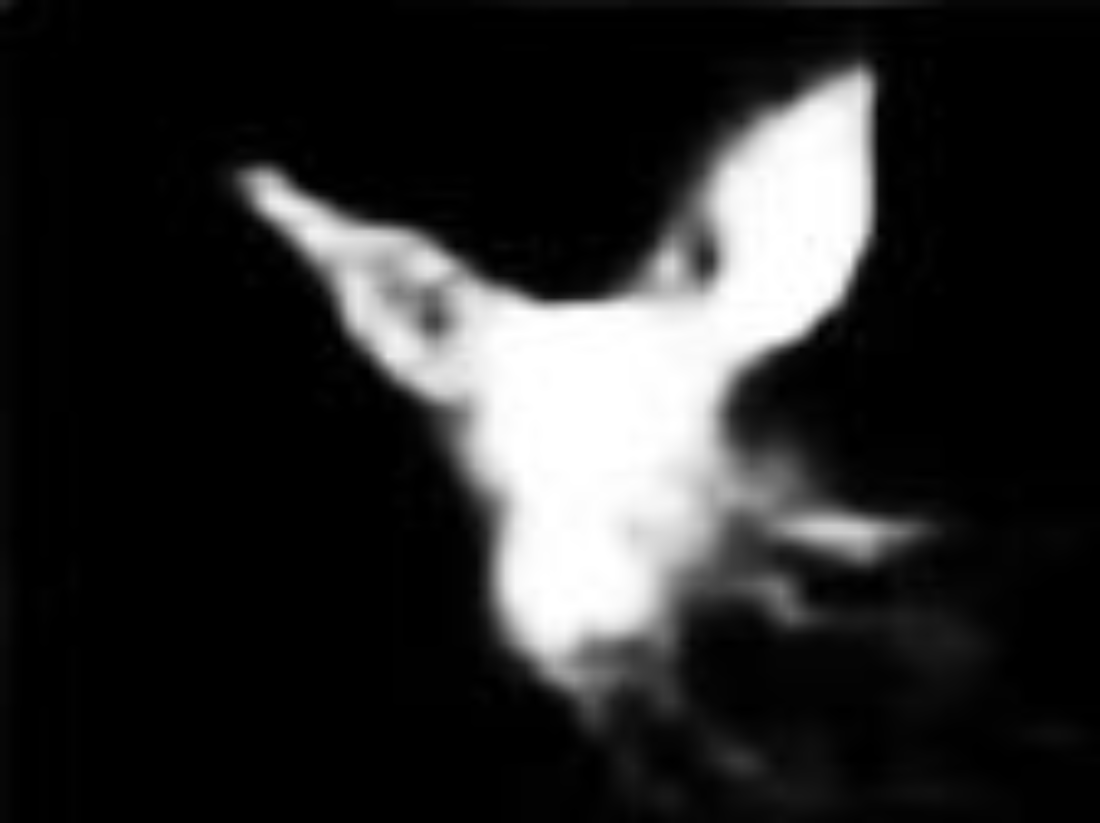}}
  \subfigure[]{\includegraphics[width=0.12\textwidth]{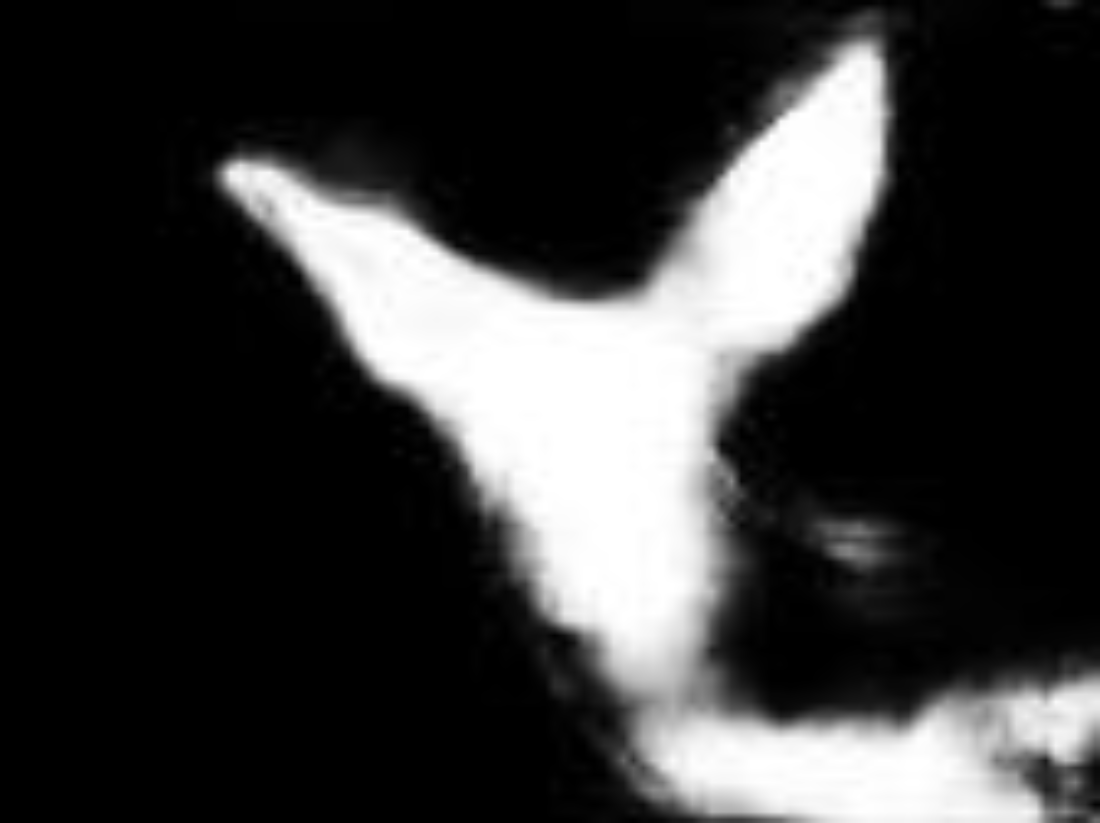}}
  \subfigure[]{\includegraphics[width=0.12\textwidth]{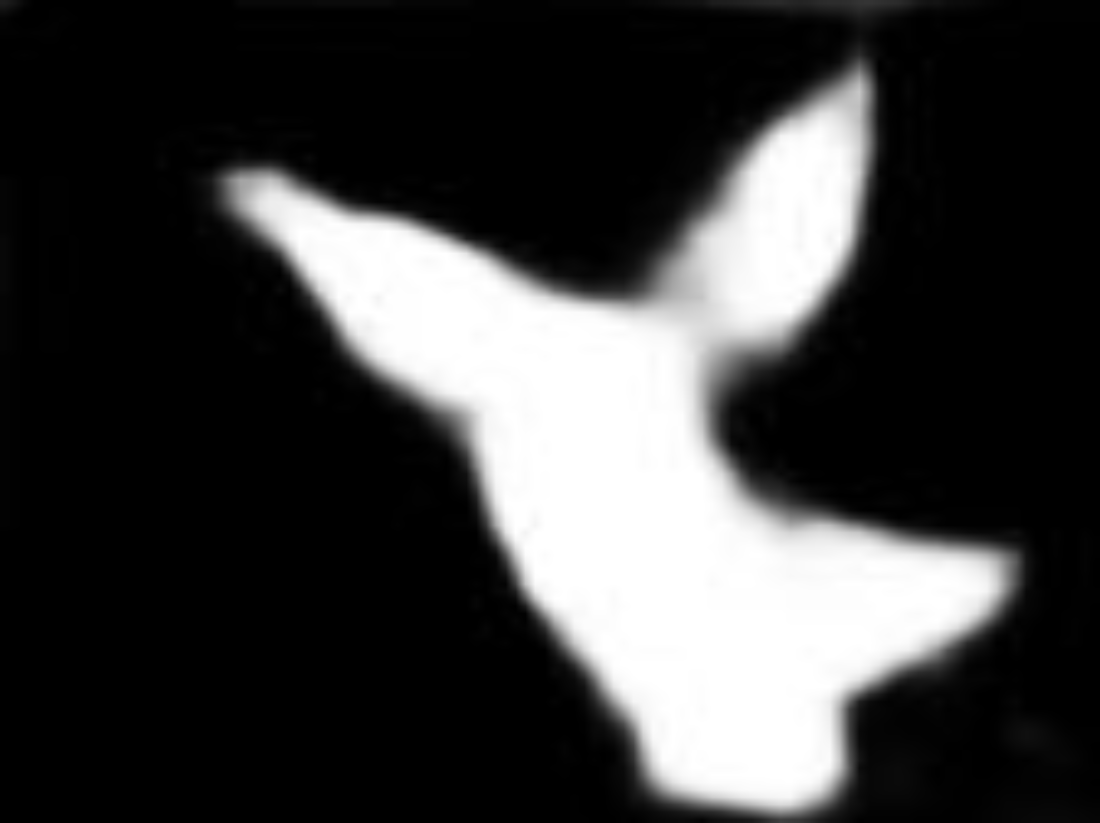}}
  \subfigure[]{\includegraphics[width=0.12\textwidth]{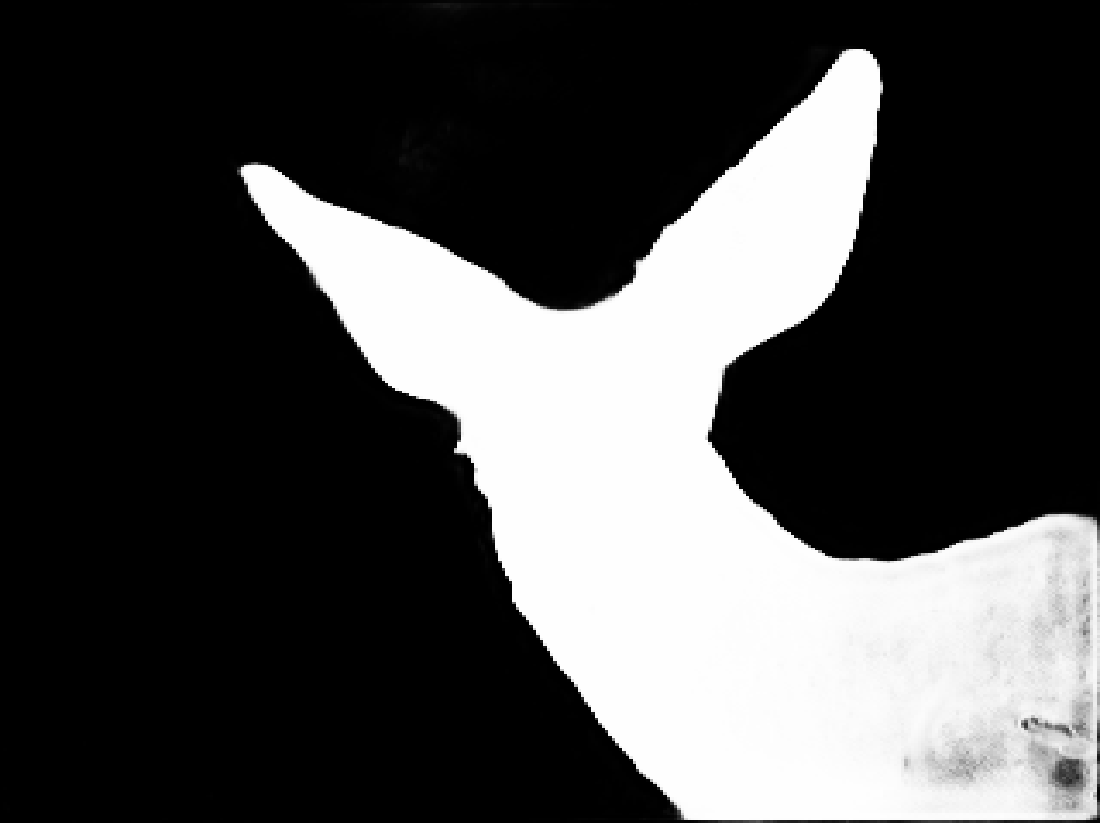}} \\
  \caption{Visual comparison of the impact of each component of our method on several examples. (a) RGB; (b) Depth; (c) Ground truth; (d) Baseline; (e) +NDAM; (f) +AIAM; (g) +JHOL; (h) +NDAM+AIAM+JHOL.}
  \label{Figure 8}
\end{figure*}

\section{Experiments}

\subsection{Experimental Setup}

\noindent \textbf{Datasets.}
We choose seven benchmark datasets including STEREO \cite{Niu2012Leveraging}, NLPR \cite{Peng2014RGBD}, RGBD135 \cite{Cheng2014Depth}, LFSD \cite{Li2014Saliency}, NJU2K \cite{Ju2015Depth-aware}, DUT-RGBD \cite{Piao2019Depth-induced} and SIP \cite{Fan2019Rethinking} as the experimental material. STEREO, also known as SSB1000, contains 1000 pairs of stereoscopic images gathered from the Internet. NLPR contains 1000 images taken under different illumination conditions. RGBD135 is also called DES, which contains 135 images about some indoor scenarios. LFSD is relatively small and contains 100 images. NJU2K contains 1985 images collected from the Internet, 3D movies and photographs. DUT-RGBD contains 1200 images taken in varied real-life situations. SIP is relatively new, which contains 929 human images. Following \cite{Pang2020Hierarchical}, we use 700 samples from NLPR, 1485 samples from NJU2K, and 800 samples from DUT-RGBD as the training set. The remaining samples and other datasets are used as the testing set.

\noindent \textbf{Evaluation Metrics.}
We employ six typical evaluation metrics including S-measure ($S_{\alpha}$) \cite{Fan2017Structure-measure}, maximum F-measure ($F^{max}_{\beta}$) \cite{Achanta2009Frequency-tuned}, average F-measure ($F^{avg}_{\beta}$) \cite{Achanta2009Frequency-tuned}, weighted F-measure ($F^{\omega}_{\beta}$) \cite{Margolin2014How}, E-measure ($E_{\xi}$) \cite{Fan2018Enhanced-alignment} and mean absolute error ($\mathcal{M}$) \cite{Perazzi2012Saliency} to comprehensively evaluate the performance of competitors. In addition, we also plot the precision-recall (PR) curves and F-measure (F$_{\beta}$) curves.

\noindent \textbf{Implementation Details.}
We implement our method based on the PyTorch toolbox with a single GeForce RTX 2080 Ti GPU.
The VGG-16 \cite{Simonyan2014Very} is adopted as the backbone. For each input image, it is simply resized to 320$\times$320 and then fed into the network to obtain prediction without any other pre-processing (e.g., HHA \cite{Gupta2014Learning}) or post-processing (e.g., CRF \cite{Krahenbuhl2011Efficient}). To avoid over-fitting, the techniques of flipping, cropping and rotation act as data augmentation. The stochastic gradient descent (SGD) optimizer is used with the batch size of 4, the momentum of 0.9 and the weight decay of 5e-4. The whole network is stopped after 30 epochs.

\subsection{Comparison with State-of-the-arts}

The proposed method is compared with other 12 state-of-the-art approaches including three RGB approaches (i.e., EGNet \cite{Zhao2019EGNet}, CPD \cite{Wu2019Cascaded} and PoolNet \cite{Liu2019A}) and nine RGB-D approaches (i.e., DF \cite{Qu2017RGBD}, CTMF \cite{Han2017CNNs-based}, MMCI \cite{Chen2019Multi-modal}, TANet \cite{Chen2019Three-stream}, AFNet \cite{Wang2019Adaptive}, DMRA \cite{Piao2019Depth-induced}, CPFP \cite{Zhao2019Contrast}, D$^{3}$Net \cite{Fan2019Rethinking} and S$^{2}$MA \cite{Liu2020Learning}). For fair comparisons, all saliency maps of these methods are provided by the authors or computed by their released codes with default settings.

\noindent \textbf{Quantitative Comparison.}
The quantitative comparison results are reported in Table \ref{Table 1}. It can be seen that our method performs best in almost all cases. For an intuitive comparison, the PR curves and F$_{\beta}$ curves are shown in Fig. \ref{Figure 5} and Fig. \ref{Figure 6}, respectively. Obviously, the curves generated by our method are closer to the top and straighter in a large range than others, which reflects its excellence and stability.

\noindent \textbf{Qualitative Comparison.}
The qualitative comparison results as shown in Fig. \ref{Figure 7}. It can be observed that our method can handle a wide variety of challenging scenes, such as blurred foreground, cluttered background, low contrast and multiple objects. More specifically, our method yields clear foreground, clean background, complete structure and sharp boundary. These results prove that our method is able to utilize cross-modal complementary information, which can not only achieve the reinforcement from the reliable depth maps but also prevent the contamination from the unreliable depth maps.

\subsection{Ablation Studies}

A series of ablation studies are conducted to investigate the impact of each core component of our method. The ablation experiment results are reported in Table \ref{Table 2}. The baseline, corresponding to scheme No. 1 (i.e., the 1st rows), refers to the network like FPNs \cite{Lin2017Feature}. It should be pointed out that the uniform evaluation metrics are redefined as a weighted sum of scores according to the proportion of each dataset in all datasets.
The visual comparison results are shown in Fig. \ref{Figure 8}.

\noindent \textbf{Effect of NDAM.}
Actually, NDAM consists of two parts, which are $P_{1}$ and $P_{2}$. By adding them into the baseline in the individual and collective manner, corresponding to scheme No. 2-3 (i.e., the 2nd and 3rd rows) and scheme No. 4 (i.e., the 4th rows), the performance is well improved. This confirms that NDAM does indeed offer additional valuable information from the channel and spatial perspectives.

\noindent \textbf{Effect of AIAM.}
Similarly, AIAM consists of two parts, which are $I_{1}$ and $I_{1}$. By adding them into the baseline in the individual and collective manner, corresponding to scheme No. 5-6 (i.e., the 5th and 6th rows) and scheme No. 7 (i.e., the 7th rows), the performance is greatly improved. This reveals the advantage of feature aggregation of AIAM.

\noindent \textbf{Effect of JHOL.}
Also similarly, JHOL (i.e., $\mathcal{L}_{JHO}$) consists of four parts, which are $\mathcal{L}_{1}$, $\mathcal{L}_{2}$, $\mathcal{L}_{3}$ and $\mathcal{L}_{4}$. By adding them into the baseline in the individual and collective manner, corresponding to scheme No. 8-11 (i.e., the 8th, 9th, 10th and 11th rows) and scheme No. 12 (i.e., the 12th rows), the performance is significantly improved. In particular, when these parts are collectively added into the baseline, there are 1.4\%, 3.6\%, 5.6\%, 2.2\%, 0.8\% and 1.4\% improvement in terms of uniform evaluation metrics in order, respectively. This suggests that the use of JHOL is crucial for the task.

\section{Conclusion}

In this paper, we propose the multi-modal and multi-scale refined network named M$^{2}$RNet for detecting salient objects. To start with, we present the nested dual attention module, which boosts the fusion of multi-modal features via the phased channel and spatial attention. Next, we present the adjacent interactive aggregation module, which boosts the aggregation of multi-scale features in the form of progressive and jumping connections. Last but not least, we present the joint hybrid optimization loss, which alleviates the imbalance of pixels from both local and global aspects. Exhaustive experimental results demonstrate the superiority of our method over the other 12 state-of-the-art approaches.

\section*{CRediT authorship contribution statement}

\textbf{Xian Fang:}
Conceptualization, Methodology, Validation, Formal analysis, Investigation, Writing - original draft, Visualization.
\textbf{Jinchao Zhu}:
Data Curation, Writing - review \& editing, Visualization.
\textbf{Ruixun Zhang}:
Writing - review \& editing.
\textbf{Xiuli Shao}:
Writing - review \& editing.
\textbf{Hongpeng Wang}:
Writing - review \& editing, Funding acquisition.

\section*{Declaration of competing interest}

The authors declare that they have no known competing financial interests or personal relationships that could have appeared to influence the work reported in this paper.

\section*{Acknowledgement}

This research was supported by the National Key R\&D Program of China under Grant 2019YFB1311804, the National Natural Science Foundation of China under Grant 61973173, 91848108 and 91848203, and the Technology Research and Development Program of Tianjin under Grant 18ZXZNGX00340 and 20YFZCSY00830.

\section*{References}

\bibliography{BibTex}

\end{document}